\pdfoutput=1
\documentclass{article}
\usepackage{iclr2027_conference,times}

\usepackage{amsmath,amsfonts,bm}

\def\eqref#1{equation~\ref{#1}}

\def\1{\bm{1}}

\DeclareMathAlphabet{\mathsfit}{\encodingdefault}{\sfdefault}{m}{sl}
\SetMathAlphabet{\mathsfit}{bold}{\encodingdefault}{\sfdefault}{bx}{n}

\usepackage{hyperref}
\hypersetup{pdftitle={CreditFilter: Calibrated Token Weights over the Within-Response Entropy Rank for RLVR}}
\usepackage{url}
\usepackage{booktabs}
\usepackage{array}
\newcolumntype{L}[1]{>{\raggedright\arraybackslash}p{#1}}
\usepackage{placeins}
\usepackage{float}
\usepackage{graphicx}
\usepackage{amsmath,amssymb}
\usepackage{amsthm}
\theoremstyle{plain}
\newtheorem{proposition}{Proposition}
\newtheorem{lemma}[proposition]{Lemma}

\newtheorem{corollary}[proposition]{Corollary}
\newtheorem{assumption}[proposition]{Assumption}
\theoremstyle{remark}

\usepackage{algorithm}
\usepackage{algpseudocode}
\usepackage{xcolor}
\usepackage{colortbl}
\definecolor{cfref}{RGB}{200,220,245}\definecolor{cfrow}{RGB}{230,239,250}\definecolor{dapref}{RGB}{250,220,185}\definecolor{daprow}{RGB}{253,239,222}\definecolor{maskref}{RGB}{200,232,205}\definecolor{maskrow}{RGB}{231,245,233}
\usepackage{multirow}
\newcommand{\method}{CreditFilter}
\newcommand{\lamPos}{0.111}
\newcommand{\lamNeg}{0.155}
\newcommand{\lamW}{0.10}
\newcommand{\lamWn}{0.14}
\newcommand{\epsW}{0.1}
\newcommand{\Hlo}{1.5}
\newcommand{\Hhi}{3.0}
\newcommand{\MainTableRows}{%
\method{} & 0.876 & 0.882 & 0.338 & 0.352 & \textbf{0.277} & \textbf{0.283} & 0.3 & 4.0 & 0.8 \\
\method{}-D & \textbf{0.887} & \textbf{0.889} & \textbf{0.367} & \textbf{0.372} & 0.268 & 0.276 & 0.7 & 5.0 & 1.9 \\
\midrule
GRPO & 0.833 & 0.848 & 0.263 & 0.281 & 0.184 & 0.173 & 1.0 & 2.4 & 4.5 \\
DAPO & 0.830 & 0.837 & 0.244 & 0.259 & 0.177 & 0.195 & 0.9 & 4.6 & 1.5 \\
80/20 mask & 0.827 & 0.817 & 0.302 & 0.294 & 0.236 & 0.227 & 3.4 & 10.9 & 7.3 \\
SC-GRPO & 0.838 & 0.836 & 0.269 & 0.283 & 0.168 & 0.156 & 1.3 & 1.1 & 4.8 \\
STARE & 0.824 & 0.819 & 0.189 & 0.183 & 0.156 & 0.155 & 0.4 & 1.3 & 1.1 \\
base model & \multicolumn{2}{c}{0.507} & \multicolumn{2}{c}{0.055} & \multicolumn{2}{c}{0.047} & 10.6 & 22.5 & 14.8 \\
}

\newcommand{\TruncRows}{%
\method{} & 6{,}346 & 0.3 & 3.6 & 9{,}797 & 4.0 & 3.5 & 9{,}188 & 0.8 & 3.3 \\
top 20\% window & 8{,}467 & 3.3 & 15.4 & 12{,}109 & 10.2 & 4.0 & 11{,}250 & 13.4 & 3.5 \\
80/20 rule, per-response cut & 6{,}228 & 5.7 & 13.3 & 10{,}972 & 25.5 & 4.8 & 9{,}699 & 20.8 & 4.6 \\
80/20 rule, batch cut & 2{,}446 & 5.8 & 0.9 & 3{,}961 & 6.6 & 0.6 & 3{,}112 & 5.6 & 0.3 \\
80/20 mask & 3{,}989 & 3.4 & 3.7 & 8{,}866 & 10.9 & 5.5 & 7{,}720 & 7.3 & 6.0 \\
SC-GRPO & 2{,}430 & 1.3 & 0.4 & 5{,}215 & 1.1 & 0.4 & 4{,}954 & 4.8 & 0.1 \\
STARE & 867 & 0.4 & 0.7 & 1{,}480 & 1.3 & 0.5 & 1{,}169 & 1.1 & 0.3 \\
DAPO & 2{,}848 & 0.9 & 7.5 & 5{,}361 & 4.6 & 11.6 & 3{,}667 & 1.5 & 7.8 \\
GRPO & 2{,}020 & 1.0 & 0.1 & 4{,}553 & 2.4 & 0.0 & 4{,}266 & 4.5 & 0.3 \\
base model (step 0) & 3{,}347 & 10.6 & 3.8 & 6{,}225 & 22.5 & 5.0 & 4{,}581 & 14.8 & 4.4 \\
}

\newcommand{\AblateRows}{%
\rowcolor{cfref} \method{} & 0.876 &  & 0.882 & 0.338 &  & 0.352 & 0.277 &  & 0.283 & 0.3 & 4.0 & 0.8 \\
\rowcolor{dapref} DAPO & 0.830 &  & 0.837 & 0.244 &  & 0.259 & 0.177 &  & 0.195 & 0.9 & 4.6 & 1.5 \\
\rowcolor{maskref} 80/20 rule, batch cut & 0.741 &  & 0.731 & 0.193 &  & 0.177 & 0.149 &  & 0.144 & 5.8 & 6.6 & 5.6 \\
\midrule
\multicolumn{13}{l}{\emph{Window shape}} \\
\rowcolor{cfrow} top 20\% window & 0.854 & {\tiny\color{gray}$-$2.2} & 0.858 & 0.380 & {\tiny\color{gray}+4.2} & 0.400 & 0.280 & {\tiny\color{gray}+0.3} & 0.274 & 3.3 & 10.2 & 13.4 \\
\rowcolor{cfrow} wide window (0.20/0.28) & 0.722 & {\tiny\color{gray}$-$15.4} & 0.750 & 0.218 & {\tiny\color{gray}$-$12.0} & 0.235 & 0.190 & {\tiny\color{gray}$-$8.7} & 0.197 & 8.8 & 9.5 & 8.6 \\
\midrule
\multicolumn{13}{l}{\emph{Coordinate}} \\
\rowcolor{cfrow} random rank & 0.849 & {\tiny\color{gray}$-$2.7} & 0.849 & 0.255 & {\tiny\color{gray}$-$8.3} & 0.248 & 0.217 & {\tiny\color{gray}$-$6.0} & 0.220 & 0.9 & 5.2 & 2.6 \\
\rowcolor{cfrow} reversed rank & 0.694 & {\tiny\color{gray}$-$18.2} & 0.703 & 0.176 & {\tiny\color{gray}$-$16.2} & 0.171 & 0.119 & {\tiny\color{gray}$-$15.8} & 0.119 & 5.4 & 11.1 & 5.7 \\
\rowcolor{cfrow} batch-level rank & 0.826 & {\tiny\color{gray}$-$5.0} & 0.833 & 0.264 & {\tiny\color{gray}$-$7.4} & 0.294 & 0.194 & {\tiny\color{gray}$-$8.3} & 0.211 & 3.1 & 6.4 & 2.6 \\
\rowcolor{maskrow} 80/20 rule, per-response cut & 0.825 & {\tiny\color{gray}+8.4} & 0.834 & 0.322 & {\tiny\color{gray}+12.9} & 0.314 & 0.263 & {\tiny\color{gray}+11.4} & 0.241 & 5.7 & 25.5 & 20.8 \\
\midrule
\multicolumn{13}{l}{\emph{Class mass}} \\
\rowcolor{cfrow} no floor & 0.862 & {\tiny\color{gray}$-$1.4} & 0.863 & 0.338 & {\tiny\color{gray}+0.0} & 0.333 & 0.254 & {\tiny\color{gray}$-$2.3} & 0.256 & 1.1 & 4.0 & 4.2 \\
\rowcolor{cfrow} rescaled masses & 0.840 & {\tiny\color{gray}$-$3.6} & 0.844 & 0.331 & {\tiny\color{gray}$-$0.7} & 0.328 & 0.243 & {\tiny\color{gray}$-$3.4} & 0.260 & 2.1 & 5.6 & 2.6 \\
\midrule
\multicolumn{13}{l}{\emph{Stabilizers}} \\
\rowcolor{cfrow} no fallback & 0.842 & {\tiny\color{gray}$-$3.4} & 0.852 & 0.327 & {\tiny\color{gray}$-$1.1} & 0.315 & 0.233 & {\tiny\color{gray}$-$4.4} & 0.233 & 2.6 & 4.7 & 3.8 \\
\rowcolor{cfrow} no damping & 0.792 & {\tiny\color{gray}$-$8.4} & 0.765 & 0.304 & {\tiny\color{gray}$-$3.4} & 0.278 & 0.228 & {\tiny\color{gray}$-$4.9} & 0.220 & 11.4 & 15.8 & 13.6 \\
\rowcolor{daprow} damping only & 0.713 & {\tiny\color{gray}$-$11.7} & 0.741 & 0.168 & {\tiny\color{gray}$-$7.6} & 0.190 & 0.138 & {\tiny\color{gray}$-$3.9} & 0.136 & 3.2 & 3.2 & 2.0 \\
}

\newcommand{\DispFullRows}{%
step 250 (in band) & 1.90 & window kept & 0.284 & 22.1 [20.2, 24.0] & 22.5 & +0.017 & 5.4 / 28.4 \\
step 250 (in band) & 1.90 & uniform & 0.535 & 11.7 [10.7, 12.6] & 12.4 & +0.077 & 3.3 / 14.8 \\
step 275 (excursion onset) & 3.24 & window kept & 0.264 & 23.1 [21.3, 24.8] & 23.3 & $-$0.085 & 4.9 / 29.4 \\
step 275 (excursion onset) & 3.24 & uniform & 0.383 & 16.1 [14.8, 17.3] & 16.8 & $-$0.016 & 3.5 / 20.4 \\
step 275 (excursion onset) & 3.24 & fallback $m_{\mathrm{gov}}\approx0.5$ & 0.316 & 20.4 [18.7, 21.9] & 20.7 & +0.023 & 4.2 / 25.8 \\
step 300 (late excursion) & 3.25 & window kept & 0.232 & 152.9 [137.6, 167.9] & 148.2 & $-$0.102 & 22.5 / 188.7 \\
step 300 (late excursion) & 3.25 & uniform & 0.234 & 67.1 [60.9, 73.2] & 63.9 & $-$0.232 & 10.6 / 82.6 \\
DAPO step 300 (control) & 4.10 & window kept & 0.198 & 56.7 [53.9, 59.2] & 55.3 & $-$0.219 & 36.4 / 64.4 \\
DAPO step 300 (control) & 4.10 & uniform & 0.250 & 65.9 [62.7, 68.7] & 64.5 & $-$0.248 & 42.2 / 74.9 \\
}

\newcommand{\HorizonRows}{%
correct & all positions (paper) & 7056 & 0.111 [0.098, 0.124] & 0.125 [0.069, 0.221] & 2.12 [1.21, 3.13] & 2.9 [1.7, 5.6] & 0.67 [0.24, 3.55] \\
correct & no truncated continuation & 4577 & 0.109 [0.095, 0.121] & 0.087 [0.061, 0.240] & 3.15 [1.91, 4.49] & 2.1 [1.2, 3.6] & 0.69 [0.29, 2.81] \\
correct & truncated graded 1 & 7056 & 0.111 [0.098, 0.124] & 0.136 [0.080, 0.226] & 2.21 [1.27, 3.21] & 2.8 [1.6, 5.0] & 0.70 [0.29, 3.53] \\
wrong & all positions (paper) & 16128 & 0.155 [0.146, 0.165] & 0.146 [0.093, 0.344] & 0.70 [0.46, 0.96] & 2.4 [1.3, 3.9] & 1.15 [0.42, 6.63] \\
wrong & no truncated continuation & 8443 & 0.150 [0.141, 0.161] & 0.110 [0.074, 0.823] & 0.66 [0.29, 1.08] & 3.4 [1.5, 7.7] & 1.05 [0.28, 10.84] \\
wrong & truncated graded 1 & 16128 & 0.155 [0.146, 0.165] & 0.164 [0.089, 0.665] & 0.51 [0.22, 0.79] & 3.0 [1.3, 6.8] & 1.20 [0.26, 12.70] \\
}

\newcommand{\FineRows}{%
correct & 0--0.0025 & 460 & 0.64 & 0.869 & 8.21 [6.04, 10.86] & 9.45 [6.90, 12.61] & 11.45 \\
correct & 0.0025--0.005 & 392 & 0.72 & 0.832 & 3.05 [2.15, 4.15] & 3.67 [2.58, 4.99] & 4.15 \\
correct & 0.005--0.01 & 408 & 0.76 & 0.796 & 2.55 [1.73, 3.48] & 3.20 [2.17, 4.38] & 3.12 \\
correct & 0.01--0.015 & 408 & 0.80 & 0.763 & 1.68 [0.98, 2.49] & 2.20 [1.27, 3.28] & 2.40 \\
correct & 0.015--0.025 & 407 & 0.83 & 0.720 & 2.82 [1.26, 5.04] & 3.92 [1.78, 6.95] & 3.87 \\
correct & 0.025--0.0375 & 401 & 0.87 & 0.671 & 0.62 [0.26, 1.03] & 0.93 [0.38, 1.53] & 0.78 \\
correct & 0.0375--0.05 & 405 & 0.89 & 0.626 & 0.80 [0.43, 1.21] & 1.28 [0.68, 1.96] & 1.27 \\
correct & 0.05--0.075 & 405 & 0.91 & 0.563 & 1.44 [0.82, 2.23] & 2.56 [1.45, 3.99] & 2.53 \\
correct & 0.075--0.1 & 401 & 0.93 & 0.485 & 0.52 [0.13, 1.08] & 1.07 [0.27, 2.24] & 0.93 \\
correct & 0.1--0.15 & 401 & 0.95 & 0.384 & 0.62 [0.19, 1.22] & 1.60 [0.48, 3.20] & 1.38 \\
correct & 0.15--0.2 & 401 & 0.96 & 0.279 & 0.41 [0.17, 0.71] & 1.47 [0.61, 2.56] & 1.43 \\
correct & 0.2--0.3 & 400 & 0.98 & 0.169 & 0.40 [0.18, 0.67] & 2.36 [1.11, 3.96] & 2.05 \\
correct & 0.3--0.4 & 400 & 0.99 & 0.082 & 0.65 [0.10, 1.63] & 7.91 [1.26, 19.31] & 8.55 \\
correct & 0.4--0.6 & 400 & 1.00 & 0.027 & 0.07 [0.00, 0.15] & 2.44 [$-$0.01, 5.31] & 2.37 \\
correct & 0.6--0.8 & 400 & 1.00 & 0.004 & 0.00 [0.00, 0.00] & 0.36 [$-$0.73, 1.26] & 0.30 \\
correct & 0.8--1 & 400 & 1.00 & 0.000 & 0.00 [0.00, 0.00] & 1.93 [0.07, 5.20] & 3.01 \\
wrong & 0--0.0025 & 534 & 0.41 & 0.943 & 1.84 [1.17, 2.73] & 1.95 [1.24, 2.91] & 2.23 \\
wrong & 0.0025--0.005 & 411 & 0.48 & 0.921 & 0.76 [0.30, 1.38] & 0.82 [0.33, 1.50] & 0.81 \\
wrong & 0.005--0.01 & 422 & 0.56 & 0.897 & 0.55 [0.30, 0.88] & 0.62 [0.33, 0.98] & 0.60 \\
wrong & 0.01--0.015 & 410 & 0.60 & 0.872 & 0.86 [0.34, 1.74] & 0.99 [0.38, 1.99] & 0.91 \\
wrong & 0.015--0.025 & 417 & 0.66 & 0.843 & 0.55 [0.27, 0.87] & 0.65 [0.32, 1.04] & 0.53 \\
wrong & 0.025--0.0375 & 417 & 0.70 & 0.805 & 0.05 [$-$0.08, 0.19] & 0.07 [$-$0.10, 0.24] & 0.10 \\
wrong & 0.0375--0.05 & 406 & 0.75 & 0.771 & 0.14 [$-$0.02, 0.34] & 0.18 [$-$0.03, 0.43] & $-$0.03 \\
wrong & 0.05--0.075 & 407 & 0.80 & 0.724 & 0.42 [0.09, 0.88] & 0.58 [0.13, 1.21] & 0.59 \\
wrong & 0.075--0.1 & 403 & 0.83 & 0.668 & 0.29 [0.05, 0.65] & 0.43 [0.07, 0.98] & 0.41 \\
wrong & 0.1--0.15 & 406 & 0.87 & 0.594 & 0.14 [0.04, 0.25] & 0.23 [0.07, 0.42] & 0.29 \\
wrong & 0.15--0.2 & 402 & 0.90 & 0.505 & 0.14 [0.01, 0.32] & 0.27 [0.02, 0.63] & 0.28 \\
wrong & 0.2--0.3 & 404 & 0.93 & 0.389 & 0.05 [$-$0.04, 0.20] & 0.14 [$-$0.11, 0.50] & 0.18 \\
wrong & 0.3--0.4 & 400 & 0.95 & 0.254 & 0.07 [0.02, 0.14] & 0.27 [0.07, 0.56] & 0.23 \\
wrong & 0.4--0.6 & 401 & 0.98 & 0.111 & $-$0.01 [$-$0.02, 0.01] & $-$0.07 [$-$0.24, 0.11] & $-$0.08 \\
wrong & 0.6--0.8 & 400 & 1.00 & 0.022 & 0.01 [0.00, 0.02] & 0.23 [$-$0.08, 0.89] & 0.32 \\
wrong & 0.8--1 & 400 & 1.00 & 0.002 & 0.00 [0.00, 0.01] & 1.20 [0.01, 5.42] & 2.73 \\
}

\newcommand{\RetentionRows}{%
uniform weights & 0.21 [0.12, 0.28] & 0.16 [0.10, 0.23] \\
top 20\% window & 0.40 [0.23, 0.53] & 0.51 [0.31, 0.61] \\
top $0.25\%$ of the rank only & 0.35 [0.17, 0.52] & 0.31 [0.13, 0.48] \\
\method{} window (exponential with its floor) & 0.61 [0.44, 0.69] & 0.64 [0.44, 0.70] \\
wide window ($0.20/0.28$ with its floor) & 0.57 [0.41, 0.65] & 0.48 [0.31, 0.58] \\
reversed rank (exponential in $1-q$) & 0.04 [0.02, 0.05] & 0.00 [0.00, 0.00] \\
exponential fitted to the forking value & 0.66 [0.45, 0.74] & 0.67 [0.47, 0.73] \\
\method{} window with a threefold top-$0.25\%$ gain & 0.79 [0.57, 0.84] & 0.79 [0.57, 0.83] \\
piecewise $C\kappa$ (fitted on the same estimates) & 0.93 [0.69, 0.95] & 0.91 [0.66, 0.92] \\
}

\newcommand{\TrackRows}{%
50 & 431 & 0.060 [0.057, 0.065] & 440 & 0.126 [0.112, 0.142] & 3 \\
100 & 302 & 0.100 [0.090, 0.108] & 578 & 0.218 [0.208, 0.229] & 2 \\
150 & 388 & 0.104 [0.091, 0.114] & 473 & 0.243 [0.230, 0.256] & 4 \\
200 & 391 & 0.099 [0.087, 0.112] & 498 & 0.223 [0.210, 0.236] & 1 \\
250 & 400 & 0.076 [0.059, 0.088] & 493 & 0.346 [0.320, 0.372] & 0 \\
300 & 483 & 0.053 [0.051, 0.055] & 294 & 0.327 [0.284, 0.369] & 13 \\
350$^\dagger$ & 471 & 0.058 [0.054, 0.061] & 173 & 0.082 [0.076, 0.087] & 28 \\
400 & 424 & 0.054 [0.052, 0.056] & 447 & 0.341 [0.317, 0.365] & 3 \\
}

\newcommand{\FallbackRows}{%
\method{} & 0.84 & 64 & 8 \\
\method{}-D & 0.69 & 30 & 10 \\
top 20\% window & 0.84 & 57 & 0 \\
wide window (0.20/0.28) & 0.65 & 37 & 9 \\
random rank & 0.60 & 42 & 8 \\
reversed rank & 1.00 & 100 & 0 \\
batch-level rank & 0.66 & 38 & 16 \\
no floor & 0.86 & 41 & 0 \\
rescaled masses & 0.83 & 64 & 3 \\
no fallback & off & --- & --- \\
no damping & 0.97 & 77 & 0 \\
}

\title{CreditFilter: Calibrated Token Weights over the Within-Response Entropy Rank for RLVR}

\author{\makebox[\dimexpr\textwidth-2\tabcolsep\relax][c]{Hanqing Zhao$^{1}$ Liuyue Xie$^{2}$ Rong-Cheng Tu$^{1}$ Xu Zhang$^{2}$ Davide Modolo$^{2}$ Dacheng Tao$^{1}$} \\
\makebox[\dimexpr\textwidth-2\tabcolsep\relax][c]{$^{1}$Nanyang Technological University \qquad $^{2}$Amazon AGI}}
\iclrfinalcopy

\begin{document}
\maketitle
\lhead{Preprint. Under review.}

\begin{abstract}

Critic-free reinforcement learning with verifiable rewards (RLVR) typically assigns a uniform advantage to every token. Token-reweighting heuristics like the 80/20 mask favor high-entropy tokens, but fail to measure a token's actual impact on the outcome. We formalize this impact as the forking value: the variance of the success probability over the policy's next-token choices. The expected forking values along a response sum exactly to the reward variance. Token credit is thus, in expectation over a prompt's responses, a fixed budget spent at the forks. This motivates weights that redistribute credit within a response. The forking value factorizes into a fork consequence, which requires costly continuations to estimate, and a fork probability, which the softmax provides. On Qwen3-8B-Base, this fork probability decays exponentially with a token's within-response entropy rank at class-specific rates, revealing that most tokens carry negligible forking value. \method{} weights each token by an exponential window over the same rank. The window is calibrated to this fork-probability envelope, under an approximately flat fork consequence. \method{} needs no extra forward passes and is stabilized by difficulty damping and an entropy-gated fallback. Averaged over three seeds, it outperforms DAPO by 7.7 points on MATH500 and by 10.5 and 9.1 points on AIME24 and AIME25. Its dynamically adaptive variant matches it. The ablations show that both parts matter: ranking tokens within each response rather than across the batch, and matching the window's decay rate to the envelope. Code is available at \url{https://github.com/yoctta/CreditFilter}.

\end{abstract}

\section{Introduction}

Credit assignment is the central open problem of reinforcement learning with verifiable rewards (RLVR)~\citep{lambert2024tulu3,guo2025deepseekr1}:
a scalar reward per response has to be spread over thousands of tokens. Per-token advantages can come from a learned value
function~\citep{schulman2016gae,schulman2017ppo}, from rollouts or perturbations at every prefix~\citep{kazemnejad2024vineppo,oar2026}, or from a
process reward model~\citep{lightman2024verify,cui2025prime}. However, each of these approaches costs a critic or extra generations. In practice, they have largely given way to the
critic-free GRPO family~\citep{shao2024deepseekmath,yu2025dapo}, which assigns a uniform advantage to every token in a response. 

\begin{figure}[t]\centering
\includegraphics[width=\linewidth]{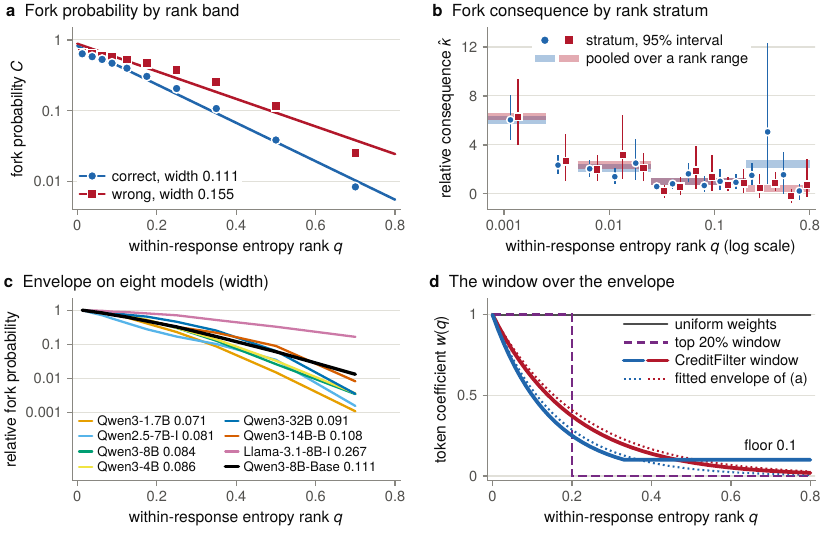}
\caption{The forking spectrum of Qwen3-8B-Base and the window built from it. (a) Fork probability $C$ in bands of the within-response entropy rank, with the exponential fits; their widths are $\lamPos$ for correct and $\lamNeg$ for wrong responses. (b) Fork consequence $\hat\kappa$ in finer rank strata on a log rank axis, relative to its class's level over ranks $2.5$--$20\%$; vertical lines: 95\% intervals from a bootstrap over prompts; light bars: values pooled over four rank ranges. (c) The envelope on eight models, each normalized to its top band; the legend gives each model's width fitted to both classes together, and for Qwen3-8B-Base the width of its correct class. (d) Uniform weights, the within-response top 20\% window and the \method{} windows, with widths $\lamW$/$\lamWn$ and floor $\epsW$ on correct responses, against the fitted envelopes of (a), drawn dotted.}
\label{fig:main}
\end{figure}

Token-importance weighting keeps this shared advantage and rescales it token by token; the 80/20 mask of \citet{wang2025_8020}, for example,
keeps only the 20\% highest-entropy tokens of a batch. The rescaling coefficient is non-negative, so every token preserves the sign
of its response's overall advantage. The underlying heuristic is that the tokens capable of changing the course of the reasoning, the \emph{forking}
tokens~\citep{bigelow2024forking}, deserve more weight, with entropy serving as a proxy for this forking potential. With the current rescaling approaches, two
questions remain open. \textbf{Which tokens are valuable forks, and how should they be weighted?} Existing rules that avoid extra forward passes rely on proxies; none measures how much changing a token shifts the probability of a correct answer. So neither the shape of the weighting function nor its cutoff is grounded in measurement.
\textbf{Ranked against which tokens?} Such masks rank tokens across the responses of a batch, so a long, uncertain and incorrect response keeps most of its tokens while a confident, correct response can lose all of them.

We answer the first question by measurement. For the second we choose the within-response rank: a theoretical identity motivates this choice, and an ablation supports it (Section~\ref{sec:ablations}). Our starting point is the \emph{forking value} $\nu$: the variance, over the policy's next-token choices, of the probability of ending with a correct answer. Along a response, \textbf{the expected forking values sum exactly to the reward variance} $V(1-V)$, where $V$ is the prompt's success probability. The identity fixes this total in expectation over a prompt's responses, not for each response. We take it as the motivation for a token weight that redistributes credit \emph{within} the response, whereas ranking tokens across a batch moves credit \emph{between} responses.

To decide where the credit goes, we factorize the forking value as $\nu=C\kappa$. The \emph{fork consequence} $\kappa$ needs expensive continuations to measure (Section~\ref{sec:spectrum}). The \emph{fork probability} $C$, the chance that two independent draws of the next token differ, comes directly from the softmax. On Qwen3-8B-Base~\citep{yang2025qwen3}, this fork probability \textbf{decays exponentially in the token's entropy rank within its response}, with separate widths for correct and wrong responses (Figure~\ref{fig:main}a). We call this profile the \emph{envelope}.

\method{} turns this envelope into a soft window (Figure~\ref{fig:main}d): it scales each token's advantage by
\begin{equation*}
w_c(q)=\max\!\big(2^{-q/\lambda_c},\,\varepsilon_c\big).
\end{equation*}
Here $q$ is the token's within-response entropy rank, and $\lambda_c$ is the envelope width of its class $c$, correct or wrong. A floor $\varepsilon_c$, applied to correct responses only, equalizes the mean weight of the two classes. The window suppresses the low-entropy tail, so tokens with a vanishing fork probability receive negligible credit.
Our contributions are:
\begin{itemize}
\item \textbf{Measurement.} We prove that the expected forking values along a response sum to the reward variance, factorize the forking value as $\nu=C\kappa$, and introduce an unbiased branching estimator to measure it. On Qwen3-8B-Base we measure the \emph{forking spectrum}, the forking value as a function of the within-response entropy rank: the fork probability follows an exponential envelope, and the fork consequence is flat apart from a peak at the head of the rank, about sixfold in the top $0.25\%$ of tokens. The envelope recurs on seven other models.
\item \textbf{Method.} We prove that, under a rank look-ahead assumption, the mean-square-optimal sign-preserving token weight targets the expected forking value. \method{} approximates this target without extra forward passes: its window is calibrated to the exponential envelope of the fork probability, under an approximately flat fork consequence. Difficulty damping and an entropy-gated fallback to uniform credit stabilize training, and an adaptive variant, \method{}-D, tracks the envelope as it moves.
\item \textbf{Results.} On Qwen3-8B-Base, averaged over three data seeds, \method{} outperforms DAPO by $7.7$ points on MATH500 and by $10.5$ and $9.1$ points on AIME24 and AIME25. On the first seed it leads GRPO by $4$, $8$ and $9$ points, exceeds the 80/20 mask, trained with its released script, by $4$--$5$ points and beats STARE and SC-GRPO on all three sets. Ablations show that both the rank \emph{within} each response and the window's width $\lambda_c$, taken from the measured envelope, matter. The windowed update moves the policy up to $2.3$ times further in KL divergence than a uniform update, without moving the parameters further. At equal steps \method{} trains on $2.1$ times DAPO's tokens, and it still leads at equal training tokens (Appendix~\ref{app:cost}).
\end{itemize}

\section{Related work}\label{sec:related}

\paragraph{Per-token credit in RLVR.}
Value functions with generalized advantage estimation~\citep{schulman2016gae,schulman2017ppo}, Monte-Carlo values from extra rollouts
(VinePPO~\citep{kazemnejad2024vineppo}), counterfactual token perturbations (OAR-P~\citep{oar2026}) and process reward
models~\citep{lightman2024verify,cui2025prime} give each token its own advantage at the price of a critic, extra generations or extra forward passes. The
critic-free GRPO family~\citep{shao2024deepseekmath,yu2025dapo,guo2025deepseekr1} keeps one scalar per response, and under shared prefixes the
group advantage already acts as a Monte-Carlo process reward~\citep{sullivan2026grpoprm}. \method{} stays in this family and moves the scalar
within the response.

\paragraph{Token reweighting and entropy control.}

A second line of work rescales the shared advantage token by token on a proxy for the fork (Appendix~\ref{app:related}): token entropy~\citep{wang2025_8020,gtpo2025}, covariance~\citep{taming2026}, gradient saliency~\citep{oar2026}, mutual information~\citep{iapo2026} and a self-conditioned KL~\citep{scgrpo2026}. \method{} adopts the entropy ordering of the 80/20 mask~\citep{wang2025_8020}, but ranks tokens within each response rather than across the batch, calibrates its window on measured fork probabilities and needs no extra forward pass. Other methods control entropy through asymmetric clipping~\citep{yu2025dapo,cliplow2025}, covariance targeting~\citep{cui2025entropy} or target-entropy loops~\citep{erpo2026,stare2026}; \method{} instead falls back toward uniform credit when the batch entropy is high.

\paragraph{Measuring forking tokens.}
\citet{bigelow2024forking} identify forking tokens, and \citet{lin2024critical} locate critical tokens within incorrect solutions, by resampling continuations for individual responses. A rank-based weight instead needs the forking value as a function of the token's rank. We estimate it as the variance over the policy's next-token alternatives, averaged within rank bands (Section~\ref{sec:spectrum}).

\section{Method}\label{sec:method}
\paragraph{Setting.}
We use DAPO as the base training recipe (Appendix~\ref{app:protocol}). For each prompt, the policy samples $G$ responses. The response $i$ receives one
group-normalized advantage $A_i$ from its final reward: $+1$ for a correct and $-1$ for a wrong answer, less an overlong penalty. We write $R_i\in\{0,1\}$ for the verifier outcome. The policy maximizes the batch token mean of the clipped surrogate $\tilde A_{i,t}\,\rho_{i,t}$,
with $\rho_{i,t}$ the importance ratio. GRPO and DAPO use $\tilde A_{i,t}=A_i$. The position $t$ of the response $i$ has prefix $s_{i,t}$, the next-token
law $\pi_{i,t}$ and entropy $H_{i,t}$ in nats. Its \emph{within-response entropy rank} is
\begin{equation}
q_{i,t}=r_{i,t}/T_i\in(0,1],
\end{equation}
where $r_{i,t}$ is the descending rank of $H_{i,t}$ among the $T_i$ tokens of the same response, so $q\approx0$ marks the most
uncertain positions of the response. The class $c(i)\in\{+,-\}$ of a response is its verifier outcome, correct or wrong, and it selects the width of the window. The floor and the damping follow the sign $s(i)$ of the advantage $A_i$ instead. The sign agrees with the outcome except when the overlong penalty moves a response across the mean of its group.

\subsection{Formulation of the window}\label{sec:window}
\method{} rescales the advantage of every token by a non-negative coefficient,
\begin{equation}\label{eq:window}
\tilde A_{i,t}=w_{i,t}A_i,\qquad
w_{i,t}=g_i\Big[(1-m_{\mathrm{gov}})+m_{\mathrm{gov}}\,k_{i,t}\Big],\qquad
k_{i,t}=\max\!\big(2^{-q_{i,t}/\lambda_{c(i)}},\,\varepsilon_{s(i)}\big).
\end{equation}
The kernel $k_{i,t}$ decays exponentially with the rank and is bounded below by a floor. Its width is the rank interval over which the weight halves. For a response whose advantage has the sign of its outcome, the kernel is the class kernel $w_c(q)=\max(2^{-q/\lambda_c},\varepsilon_c)$. The factor $g_i$ damps the prompts that the policy mostly solves, and $m_{\mathrm{gov}}$ is an entropy-gated fallback to uniform credit (Section~\ref{sec:stabilizers}); with $g_i=m_{\mathrm{gov}}=1$ the window is unmodified.
Figure~\ref{fig:main}d shows this window over the fitted envelope, and Algorithm~\ref{alg:creditfilter} in Appendix~\ref{app:algorithm} gives the computation for one batch.

\paragraph{Shape and width from the envelope.}
Let $C=1-\sum_v\pi_v^2$ be the fork probability of a position. The envelope of a class is the mean of $C$ at each rank, $\mathbb E[C\mid q,c]$. We fit it with a single exponential in the rank and take the width of the class from this fit:
\begin{equation}\label{eq:calib}
\lambda_c\approx\lambda_{C,c},\qquad\text{the width of the class envelope profile }\mathbb E[C\mid q,c],
\end{equation}
where the fit is log-linear over ten rank bands covering $q\le0.8$ (Appendix~\ref{app:track}). \textbf{\boldmath\method{} uses $\lambda_+=\lamW$ and $\lambda_-=\lamWn$}, the envelope widths measured on a subset of the base-policy responses and rounded. The full measurement of Section~\ref{sec:spectrum} gives $\lamPos$ and $\lamNeg$ (Table~\ref{tab:spectrum}).
The fork probability is the factor of the forking value that the softmax provides. The forking value is
\begin{equation}\label{eq:nu}
\nu_t=\mathrm{Var}_{x\sim\pi(\cdot\mid s_t)}\,Q(s_t,x),
\end{equation}
the variance of the success probability $Q$ over the next token. It factors as $\nu=C\kappa$ (Proposition~\ref{prop:factor}), where the \emph{fork consequence} $\kappa$ needs continuations to evaluate. Section~\ref{sec:theory} shows that the \emph{forking spectrum} $\mathbb E[\nu\mid q,c]$ is, up to a value factor, the target of a non-negative coefficient (Proposition~\ref{prop:mmse}).

\paragraph{Class mass and the floor.}
Adam removes a constant scale of the gradient (Lemma~\ref{lem:adam}). While the window is fully on and gradient clipping is inactive, only the shape of the window within a class and the ratio of the two class masses $M_c=\mathbb E_q[w_c(q)]$ therefore reach the update.
The mass $M$ has a closed form in terms of $(\lambda,\varepsilon)$ (Proposition~\ref{prop:mass}). \textbf{\method{} equalizes the class masses by introducing a positive-class floor}:
\begin{equation}\label{eq:mass}
M(\lambda_+,\varepsilon_+)=M(\lambda_-,0),
\end{equation}
which has a unique solution when $\lambda_->\lambda_+$. Given the widths $\lamW/\lamWn$, setting the floor to $\varepsilon_+=\epsW$ (the solution $0.106$, rounded) brings the class masses to $0.196$ and $0.200$. The floor keeps a small positive update on the low-entropy tokens of correct responses. The wrong class has no floor, so its negative
gradient concentrates on the forking positions.

\paragraph{Why the within-response rank.} Appendix~\ref{app:coord} gives the three reasons: every response of a given length receives \textbf{the same mean coefficient}, so credit is redistributed within each response rather than between responses; the rank is \textbf{invariant to monotone transforms} of the uncertainty measure; and on the rank the exponential is \textbf{the memoryless kernel}, so a single width sets the concentration.

\subsection{Damping and fallback}\label{sec:stabilizers}
\emph{Difficulty damping} scales the responses with a positive advantage by $g_i=(1-\hat p_i)^{\gamma}$, where $\gamma=1$ and
$\hat p_i$ is the pass rate of the response's group, and leaves the other responses undamped. For uniform token weights and in expectation over a prompt's responses, the damped update preserves the direction of the undamped one and shrinks it by a factor between $3/4$ and $1$
(Proposition~\ref{prop:damp}). Because $g_i$ shrinks as $\hat p_i$ grows, the update on prompts the policy mostly solves rests mainly on the wrong responses. With the window, removing the damping costs 3--8 points (Section~\ref{sec:ablations}). As a safeguard, an \emph{entropy-gated fallback} reverts the update toward
uniform credit as the batch mean entropy $\bar H$ rises,
\begin{equation}
m_{\mathrm{gov}}=\mathrm{clip}\Big(\frac{H_{\mathrm{hi}}-\bar H}{H_{\mathrm{hi}}-H_{\mathrm{lo}}},\,0,\,1\Big),\qquad H_{\mathrm{lo}}/H_{\mathrm{hi}}=\Hlo/\Hhi\ \text{nats}.
\end{equation}
Above $H_{\mathrm{hi}}$ the window is replaced by uniform token weights, while damping and clipping stay. Over the \method{} training run of Section~\ref{sec:experiments}, the window is fully active in
64\% of the steps and fully replaced by uniform credit in 8\% (Appendix~\ref{app:arms}).

\subsection{Re-measuring the width online (\method{}-D)}\label{sec:dynamic}
The calibration of~\eqref{eq:calib} is a property of the current policy. The envelope moves during training (Section~\ref{sec:spectrum}), so widths fixed at the base policy's values become mismatched later in training.
Since $C$ comes from the same softmax as the entropies, every batch gives an estimate $\hat\lambda_{C,c,t}$ of each class width at no extra cost. \textbf{\method{}-D tracks it with an
exponential moving average}:
\begin{equation}\label{eq:track}
\lambda_{c,t}=(1-\alpha)\,\lambda_{c,t-1}+\alpha\,\hat\lambda_{C,c,t},\qquad\alpha=0.1,
\end{equation}
where the batch estimate and the average are clamped to a class-specific range, with the average initialized at the measured base widths $\lamPos/\lamNeg$ (Appendix~\ref{app:track}). With moving widths, a fixed floor can no longer equalize the classes, so \method{}-D drops the floor and, at every step, rescales the window of the class with the smaller natural mass $M(\lambda_{c,t},0)$ up to match the larger mass. This equalizes the masses and keeps the shape of each window.

\section{Theory: the target of a sign-preserving token weight}\label{sec:theory}

Three results identify the target of a sign-preserving token weight and explain why the window approximates it. Proposition~\ref{prop:credit} gives the expected total of the forking value along a response.
Proposition~\ref{prop:mmse} shows that the conditional expectation of the forking value, given an observable of the token, is the target of a non-negative token weight.
Proposition~\ref{prop:factor}
factors the forking value into the fork probability, which the softmax provides and which bounds the forking value, and the fork consequence, which needs continuations to estimate.
The proofs are in Appendix~\ref{app:theory}. The results describe the 0--1 verifier outcome under the sampling policy and rest on one assumption about the rank; the section closes with how far both hold.

\paragraph{Setting.}
Consider on-policy rollouts for a fixed prompt, with reward $R\in\{0,1\}$ the verifier outcome. At position $t$ with prefix $s_t$, the value, the action value of a next token $x$ and the credit increment of the sampled token $x_t$ are
\begin{equation*}
V_t=\mathbb E[R\mid s_t],\qquad Q(s_t,x),\qquad \delta_t=Q(s_t,x_t)-V_t.
\end{equation*}
The conditional variance of this increment is the forking value of~\eqref{eq:nu}. The rescaling gives token $t$ of response $i$ the advantage $w(o_{i,t})A_i$, where the coefficient $w\ge0$ depends on an observable $o$ of the token. Expectations over positions use position weights, which are uniform over all tokens under a token-mean loss. We write
\begin{equation*}
N(o)=\mathbb E[\delta_tA\mid o],\qquad D(o)=\mathbb E[A^2\mid o],\qquad m(o)=\mathbb E[\nu_t\mid o].
\end{equation*}

\begin{proposition}[credit decomposition]\label{prop:credit}
With increments padded by zero after termination,
\begin{equation}
\textstyle\sum_t\delta_t=R-V_1,\qquad \mathbb E[\delta_t\mid s_t]=0,\qquad \mathbb E[\delta_t\delta_u]=0\ \ (t\ne u),\qquad
\mathbb E[\delta_t^2\mid s_t]=\nu_t,
\end{equation}
and $\sum_t\mathbb E[\nu_t]=\mathrm{Var}(R)=V_1(1-V_1)$.
Thus the reward variance equals the expected sum of the forking values along a response.
\end{proposition}
The identity is the orthogonal-increment decomposition of the bounded martingale $V_t$, and it \textbf{motivates ranking tokens within each response} (Section~\ref{sec:window}). It fixes the mean of this sum over a prompt's responses, not the sum of each response (Appendix~\ref{app:coord}).

\begin{assumption}[rank look-ahead]\label{ass:lookahead}
Let $\Gamma_t=(\#\{u>t:H_u>H_t\},\,T)$, with ties broken by position (earlier first) and $T$ the number of ranked tokens, so that the within-response rank $q_t$ is a function of $(s_t,\Gamma_t)$. For every position $t$,
\begin{equation*}
\mathbb E[\delta_t\mid s_t,\Gamma_t,R]=\mathbb E[\delta_t\mid s_t,R].
\end{equation*}
Given the prefix and the outcome, the number of later tokens more uncertain than the current one and the response length carry no further information about the current increment.
\end{assumption}
The assumption is needed because the weight is a function of the rank, and the rank looks ahead: it counts later tokens and is fixed only when the response ends. Under it, the expected increment given the prefix and the outcome is not changed by also conditioning on the rank, which is the step Proposition~\ref{prop:mmse} uses.

\begin{proposition}[positive-part MMSE coefficient and its target]\label{prop:mmse}
Among $w\ge0$, the mean-square error $\mathbb E[(\delta_t-w(o)A)^2]$ is minimized on $\{D>0\}$ by
\begin{equation*}
w^*(o)=\max\{0,\,N(o)/D(o)\}.
\end{equation*}
Let $R$ be the verifier outcome, $c$ the class it defines, $A=(R-V_1)/d$ with $d$ a constant of the prompt, and let the position weights be functions of $(s_t,\Gamma_t,R)$ (token-mean weights and weights that count every rollout equally both qualify). Under Assumption~\ref{ass:lookahead}, for a fixed prompt and $o=(q_t,c)$,
\begin{equation}\label{eq:ident}
N(o)=\frac1d\,\mathbb E\big[(1-V_1)\,\nu_t/V_t\,\big|\,o\big]\ \textup{(correct)},\qquad N(o)=\frac1d\,\mathbb E\big[V_1\,\nu_t/(1-V_t)\,\big|\,o\big]\ \textup{(wrong)},
\end{equation}
and $D$ is constant within each class. Hence $w^*$ is proportional to this value-weighted forking spectrum within each class. It is proportional to the forking spectrum $\mathbb E[\nu_t\mid o]$ when the $\nu$-weighted mean of $1/V_t$ (correct) or $1/(1-V_t)$ (wrong) at rank $q$ does not depend on $q$. For the group advantage of $G$ rollouts the identities hold with the prompt-level scales $\mathbb E[A\mid\mathrm{prompt},R]$ in place of $(1-V_1)/d$ and $-V_1/d$ (Appendix~\ref{app:credit}).
\end{proposition}
\begin{proof}[Proof sketch]\renewcommand{\qedsymbol}{}
The positive part minimizes the quadratic $D w^2-2Nw$ over $w\ge0$. Since $\delta_t$ is determined by $s_{t+1}$ and $V_{t+1}=V_t+\delta_t$,
\begin{equation*}
\mathbb E[\delta_tR\mid s_t]=\mathbb E[\delta_tV_{t+1}\mid s_t]=\mathbb E[\delta_t^2\mid s_t]=\nu_t.
\end{equation*}
With $\mathbb E[\delta_t\mid s_t]=0$ this gives
\begin{equation*}
\mathbb E[\delta_t\mid s_t,R=1]=\frac{\nu_t}{V_t},\qquad \mathbb E[\delta_t\mid s_t,R=0]=-\frac{\nu_t}{1-V_t}.
\end{equation*}
Assumption~\ref{ass:lookahead} extends both to conditioning on $(s_t,\Gamma_t,R)$, of which $o$ and the position weights are functions, so the weights leave the conditional law of $\delta_t$ unchanged. On a correct response $A=(1-V_1)/d$ is fixed by the prompt, which gives the first identity; on a wrong one $A=-V_1/d$ and the two signs cancel. $D$ equals $(1-V_1)^2/d^2$ or $V_1^2/d^2$ on the two classes. The group-advantage case and the strata with $m(o)=0$ are treated in Appendix~\ref{app:credit}.
\end{proof}
\textbf{The target of the window is therefore the forking spectrum} $\mathbb E[\nu_t\mid o]$ \textbf{up to the \emph{value factor}}, the weight $1/V_t$ or $1/(1-V_t)$ on $\nu_t$ in~\eqref{eq:ident}. Our window follows the spectrum over the rank and leaves out the value factor. The criterion also determines the class masses. For a fixed prompt with equal position weights and $N\ge0$, the per-token class mean of $w^*$ is $d/\mathbb E[T\mid c]$, so classes of equal mean length receive equal masses. The class-mass equalization of Section~\ref{sec:window} imposes equal masses in all cases.

\begin{proposition}[factorization and envelope]\label{prop:factor}
At a position with next-token distribution $\pi$ and action values $Q_k\in[0,1]$,
\begin{equation}
\nu=\tfrac12\sum_{k,l}\pi_k\pi_l(Q_k-Q_l)^2=C\,\kappa,\qquad C=1-\sum_k\pi_k^2,\qquad
\kappa=\tfrac12\,\mathbb E\big[(Q_K-Q_L)^2\,\big|\,K\ne L\big],
\end{equation}
with $K,L\sim\pi$ independent, $\kappa\in[0,\tfrac12]$ and $\kappa:=0$ where $C=0$. With $H$ the entropy in nats,
\begin{equation}
\nu\ \le\ C/2\ \le\ 1-\pi_{\max}\ \le\ C\ \le\ H.
\end{equation}
\end{proposition}
The consequence $\kappa$ is half the mean squared value gap between two independent draws from $\pi$. The
upper bound $\nu\le C/2$ suppresses the low-$C$ tail of the forking spectrum: \textbf{a position with a vanishing fork probability has a vanishing forking value}, whatever its consequence. The value-weighted target inherits the bound, scaled by the value factor.
Over the rest of the rank, \textbf{weighting by the envelope is a modeling choice}. Suppose the mean fork consequence, given the rank and the fork probability, is constant within a class (condition K). Then the forking spectrum is proportional to the class envelope $\mathbb E[C\mid q,c]$, and under the same condition on the value-weighted consequence so is the target (Corollary~\ref{cor:window}). Section~\ref{sec:spectrum} measures how far condition K holds and how closely the envelope matches the forking spectrum.

\paragraph{How far the results hold.}
The propositions describe the 0--1 outcome under the policy that sampled the rollouts. Training also subtracts the overlong penalty, which lowers the reward only of responses longer than 16{,}384 tokens (Appendix~\ref{app:protocol}); it is outside the propositions. \textbf{Assumption~\ref{ass:lookahead} is an approximation.} In a regression on the base policy (Appendix~\ref{app:atwo}), the part of the increment that the assumption rules out is within sampling error when pooled over the top $2.5\%$ of the rank. In finer bands it shows only in the top $0.25\%$ of correct responses, where it has the sign of the advantage. Counting it, the top $2.5\%$ of the rank holds $28\%$ of the correct-class target instead of $25\%$: the approximation puts slightly too little weight at the head, where the window is already largest.

\section{Experiments}\label{sec:experiments}

\noindent\textbf{Setup.} All runs train Qwen3-8B-Base on the DAPO-17k prompts~\citep{dapo17k} for 400 steps with 128 prompts and 16 responses per step, mini-batch 32 and learning rate
$10^{-6}$. Responses run up to 20{,}480 tokens, with dynamic sampling, the overlong penalty and clip ratios $0.2$ below and $0.27$ above ($0.28$ above for DAPO,
symmetric $0.2$ for GRPO; every setting in Appendix~\ref{app:protocol}). We train the 80/20 mask of \citet{wang2025_8020} with its released training script on our training set and schedule, and STARE~\citep{stare2026} and SC-GRPO~\citep{scgrpo2026} with their published settings (Appendix~\ref{app:arms}). By default a run uses the data seed 201, which sets the order of the training prompts, and methods are compared between runs with the same seed. We report mean@32 (temperature $1.0$, top-$p$ $0.7$) on MATH500~\citep{hendrycks2021math,math500} ($n=500$), AIME24 and AIME25~\citep{aime2024,aime2025} ($n=30$ each) averaged over
checkpoints 350/375/400 (the late average) and at step 400.

\begin{table}[t]\centering\footnotesize
\caption{mean@32 after training Qwen3-8B-Base with data seed 201, on MATH500 ($n{=}500$), AIME24 and AIME25 ($n{=}30$ each): the average over checkpoints 350/375/400 (late), the value at step 400, and the share of step-400 samples left unanswered at the 20{,}480-token cap. Bold: best accuracy in each column; the base-model row is step 0.}
\label{tab:main}
\setlength{\tabcolsep}{4pt}\renewcommand{\arraystretch}{0.9}
\begin{tabular}{l*{6}{c}ccc}\toprule
 & \multicolumn{2}{c}{MATH500} & \multicolumn{2}{c}{AIME24} & \multicolumn{2}{c}{AIME25} & \multicolumn{3}{c}{unanswered (\%)} \\
\cmidrule(lr){2-3}\cmidrule(lr){4-5}\cmidrule(lr){6-7}\cmidrule(l){8-10}
Run & late & 400 & late & 400 & late & 400 & M500 & A24 & A25 \\ \midrule
\MainTableRows
\bottomrule
\end{tabular}
\end{table}

\subsection{The forking spectrum of Qwen3-8B-Base}\label{sec:spectrum}
At a sampled position of a base-policy response, an instrument branches on several next tokens, continues each branch using the training sampler, and evaluates the continuations. It estimates $\nu$ as the variance of the branch success probabilities under the policy, unbiased over the next tokens it branches on; Appendix~\ref{app:instrument} gives the two instruments, the protocol and the full tables.

\noindent\textbf{The fork probability decays exponentially in the rank; the consequence peaks in the top $0.25\%$ (Figure~\ref{fig:main}a,\,b).} A single exponential fits the band means of $C$ with widths $\lamPos$ (correct) and $\lamNeg$ (wrong). On the four-branch instrument (Table~\ref{tab:bands}), half of the correct class's forking value is concentrated in the top $10\%$ of ranks, with the top $20\%$ capturing $68\%$ of it ($63\%$ for wrong responses). The fork consequence is flat over most of the rank and rises only at the head (Table~\ref{tab:fine}). Compared with its mean over ranks $2.5$--$20\%$, it is \textbf{\boldmath about six times larger in the top $0.25\%$ of the rank}, which is about two tokens per response, and about twice as large over ranks $0.5$--$2.5\%$; beyond $20\%$ the estimates are too noisy to show a trend. Condition~K holds approximately: the consequence is flat over the rank apart from a spike in the top $2.5\%$, the most uncertain tokens of a response. There the envelope already gives those tokens their largest weight. Relative to the ranks just below, the spike would add a further factor of about three on a handful of tokens (the next paragraph quantifies it). We therefore approximate the consequence as uniform over the rank. The exponential envelope of the fork probability recurs across seven other models, with widths from $0.071$--$0.108$ on the Qwen models to $0.267$ on Llama-3.1-8B-Instruct, so the width is calibrated per model (Figure~\ref{fig:main}c, Table~\ref{tab:models}).

\noindent\textbf{What the window retains.} We score a weight profile by the share of the matched-filter gain that it retains. This share is its squared cosine with the measured forking spectrum over the rank, and it is $1$ for the spectrum itself. The \method{} window retains $0.61$ (correct) and $0.64$ (wrong), against $0.40$ and $0.51$ for the within-response top 20\% window and $0.21$ and $0.16$ for uniform weights (Figure~\ref{fig:main}d, Table~\ref{tab:retention}). Applying a threefold weight explicitly to the top $0.25\%$ would raise both shares to $0.79$, a gain the window leaves out (Appendix~\ref{app:fine}).

\noindent\textbf{The envelope moves during training.} On rollouts of the \method{} checkpoints sampled at the training length, the wrong-class width doubles to approximately $0.34$ from step 250 onward (with the exception of 350). Conversely, the correct-class width halves to about $0.05$ from step 300 onward (Table~\ref{tab:track}). Despite this shifting envelope, the fixed window maintains a squared cosine of at least $0.83$ with the moving envelope (Appendix~\ref{app:track}).

\subsection{Main results}\label{sec:main}
\noindent\textbf{\method{} leads DAPO on average over three seeds, and every baseline on the first seed.} Table~\ref{tab:seedmean} gives the late averages over three data seeds (Appendix~\ref{app:seeds}).
\begin{table}[t]\centering\footnotesize
\caption{Late averages over three data seeds, mean $\pm$ standard deviation. Table~\ref{tab:seeds} lists the runs of each seed.}
\label{tab:seedmean}
\begin{tabular}{lccc}\toprule
Run & MATH500 & AIME24 & AIME25 \\ \midrule
\method{} & $0.860\pm0.025$ & $0.326\pm0.042$ & $0.258\pm0.035$ \\
DAPO & $0.783\pm0.107$ & $0.221\pm0.050$ & $0.167\pm0.032$ \\
\bottomrule
\end{tabular}
\end{table}
\textbf{\boldmath The mean difference is $7.7$, $10.5$ and $9.1$ points}, and on both AIME sets \method{}'s lowest seed is above DAPO's highest. Table~\ref{tab:main} gives the first seed, on which every method was trained. There \method{} leads DAPO by 5, 9 and 10 points and GRPO by 4, 8 and 9 points; on every set the first seed's margin over DAPO is the median of the three seeds. It leads the 80/20 mask by 4--5 points, and exceeds STARE and SC-GRPO by 4--15 points. \method{}-D, which tracks the widths during training, scores higher than \method{} on MATH500 and is comparable on the AIME sets. At equal steps \method{} trains on $2.1$ times as many tokens as DAPO, because its responses are longer; Appendix~\ref{app:cost} gives the training cost and the comparisons at equal training tokens and at equal decoding budgets.

\subsection{Entropy dynamics and local displacement}\label{sec:entdyn}
\begin{figure}[t]\centering
\includegraphics[width=\linewidth]{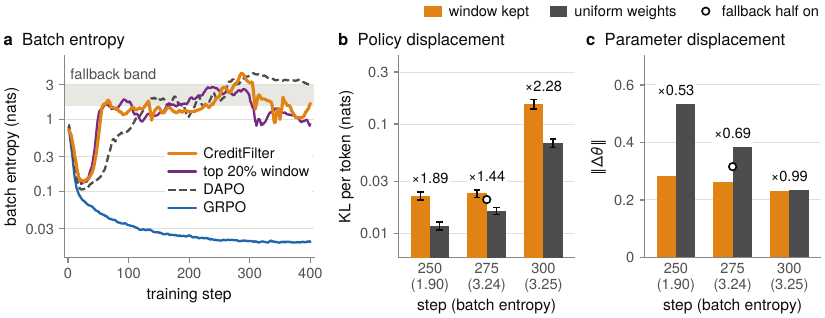}
\caption{Training dynamics and local displacement. (a) Batch entropy as a five-step moving average, with the fallback band of 1.5--3 nats shaded: GRPO collapses and DAPO stays above 3 nats after step 263. (b, c) Five-step continuations from three checkpoints of the \method{} run, each labeled with its step and batch entropy. Keeping the window moves the policy 1.4--2.3 times further in KL than uniform weights in (b), and moves the parameters, $\|\Delta\theta\|=\|\theta_{t+5}-\theta_t\|_2$, no further in (c). Numbers: window/uniform ratio; error bars: 95\% prompt-bootstrap intervals, all listed in Table~\ref{tab:disp_full}; open circle: a third continuation at step 275, with the fallback half on, $m_{\mathrm{gov}}\approx0.5$.}
\label{fig:dyn}
\end{figure}
\noindent\textbf{The windowed update moves the policy further without moving the parameters further.} The batch entropy of the nine runs plotted in Figure~\ref{fig:train} (Appendix~\ref{app:extra}) collapses to about $0.1$ nats within 25 steps. \method{}'s batch entropy then stays between $0.7$ and $2.2$ nats for two hundred steps and has one brief spike to $4.4$ nats around step 290, which resolves by step 307. To measure how far a few updates move the model, its \emph{local displacement}, we take three checkpoints of the \method{} run: step 250 (inside the fallback band), step 275 (at the onset of the entropy spike), and step 300 (during its descent). From each checkpoint we run two five-step continuations: one keeps the \method{} window with the fallback switched off, the other uses uniform weights with damping and clipping kept. For each we measure the token-level KL from the starting policy, evaluated on the starting policy's own rollouts, and the parameter displacement $\|\theta_{t+5}-\theta_t\|_2$ (Appendix~\ref{app:disp}). At every checkpoint the windowed update, with a fifth of the weight mass, moves the parameters no further than the uniform one. Yet it shifts the policy \textbf{\boldmath$1.4$--$2.3$ times further in KL} (Figure~\ref{fig:dyn}b,c; Table~\ref{tab:disp_full}). By contrast, at the step-300 checkpoint of the DAPO control run, the window/uniform KL ratio is $0.86$ (Appendix~\ref{app:disp}).

\subsection{Ablations}\label{sec:ablations}
\begin{table}[t]\centering\footnotesize
\caption{Ablations: each row changes one component of \method{}; the DAPO row repeats Table~\ref{tab:main}, and the two 80/20-rule rows apply the mask's selection rule on the settings of the DAPO row, with the cut taken over the rollout batch or within each response. Columns as in Table~\ref{tab:main}; the late columns add the difference in points from the row's reference, marked by the row shading (reference rows darker).}
\label{tab:ablate}
\scriptsize\setlength{\tabcolsep}{3pt}\renewcommand{\arraystretch}{0.92}
\begin{tabular}{l*{3}{r@{\hspace{3pt}}lc}ccc}\toprule
 & \multicolumn{3}{c}{MATH500} & \multicolumn{3}{c}{AIME24} & \multicolumn{3}{c}{AIME25} & \multicolumn{3}{c}{unanswered (\%)} \\
\cmidrule(lr){2-4}\cmidrule(lr){5-7}\cmidrule(lr){8-10}\cmidrule(l){11-13}
Run & \multicolumn{2}{c}{late} & 400 & \multicolumn{2}{c}{late} & 400 & \multicolumn{2}{c}{late} & 400 & M500 & A24 & A25 \\ \midrule
\AblateRows
\bottomrule
\end{tabular}
\end{table}
\noindent\textbf{Window and damping are complementary.} Against DAPO, the window alone (the run without damping) loses 4 points on MATH500 and gains 6 on AIME24 (Table~\ref{tab:ablate}). Damping alone loses 4--12 points, and the two together gain 5--10. We read the window as moving a correct response's update onto its forking tokens and off its execution tokens. Execution tokens are the low-entropy tokens that carry out steps already decided, including those that close the reasoning and box the answer. The pressure to terminate then weakens: reasoning grows longer, which pays off on AIME, and 11--16\% of step-400 samples reach the 20{,}480-token cap unanswered. Damping shrinks the correct-class update on training prompts the policy mostly solves, so the few wrong responses carry the update on those prompts; we expect most MATH500 problems to be of this kind. Their negative credit at the forks plausibly curbs digressions: the unanswered share falls to 0--4\%, and the AIME gains remain.

\noindent\textbf{The gain needs the entropy order and the within-response rank.} Reversing the rank collapses the entropy to $0.01$ nats, and a random within-response rank lands only 1--4 points above DAPO. Ranking across the batch stays within 2 points of DAPO, and on MATH500 it is 2 points below even the random rank. We read the reversed rank as reinforcing the most confident tokens until the policy stops exploring. A random rank still redistributes credit within each response but reaches the forks only by chance, so the entropy order supplies most of the gain. A batch-level rank moves credit between responses: long uncertain responses, mostly wrong, keep most of their tokens, and concise correct ones lose theirs. On the settings of the DAPO row, the 80/20 selection rule gains 8--13 points when its cut is taken within each response instead of over the batch.

\noindent\textbf{The exact shape matters less than the width.} The window retains $0.61/0.64$ of the matched-filter gain and the top 20\% window $0.40/0.51$ (Table~\ref{tab:retention}), yet over three seeds their accuracies are close (Table~\ref{tab:seeds}). The exponential window finishes more of its responses, with fewer tokens and at a lower training cost (Table~\ref{tab:expvstop}). Doubling the widths to $0.20/0.28$ from the first step retains $0.57/0.48$ and costs 9--15 points. Training thus seems to respond mainly to where the mass sits, and both shapes put it at the head of the rank. The top 20\% window's zero tail drops the small update on the closing tokens of correct responses, so 3--13\% of its samples reach the cap unanswered. Early on, doubled widths spread credit into middle ranks outside the base policy's envelope ($0.11/0.16$), onto execution tokens and noise.

\noindent\textbf{The floor, width tracking and the fallback each add a small gain.} Dropping the floor costs 1 and 2 points on MATH500 and AIME25; rescaling the class masses at fixed widths costs 4 on MATH500. The floor and the rescaling both equalize the class masses. The floor adds a small update in the tail of correct responses. The rescaling raises their head weights about 1.4-fold; like removing the damping, this enlarges the correct-class update. Adding the tracking to the rescaled run gives \method{}-D, 5 and 4 points higher on MATH500 and AIME24, plausibly by following the moving envelope. Without the fallback, the late accuracy is 3 and 4 points lower on MATH500 and AIME25. We hypothesize that at high batch entropy many high-entropy tokens are interchangeable wording, so the rank stops finding forks. The fallback then moves the update toward uniform credit, and the run without it keeps crediting noise.

\section{Conclusion}
In critic-free RLVR the expected forking values along a response sum to the reward variance, which motivates allocating each response's credit at its own forks. The fork probability, which bounds the forking value and comes directly from the softmax, decays exponentially in a token's within-response entropy rank at a class-specific rate. \method{} turns this envelope into a token weight with no extra forward passes. Averaged over three seeds, it gains $7.7$ points on MATH500 and $10.5$ and $9.1$ points on AIME24 and AIME25 over DAPO, at about twice DAPO's training tokens. On fixed training settings both the rank within each response and the window's width matter: ranking across the batch instead of within each response loses 5--8 points, and doubling the window's widths loses 9--15.

\paragraph{Limitations.}
All training results are on one base model and one prompt set, with 30-problem AIME sets and a 20{,}480-token cap. The ablations and the reproduced baselines are single runs. \method{}'s responses are longer than the baselines', and at equal steps it trains on about twice DAPO's tokens (Appendix~\ref{app:cost}). The theory is tested on the base policy only: Assumption~\ref{ass:lookahead} and the flat fork consequence hold approximately, the window leaves out the value factor, and the propositions do not cover the overlong penalty (Section~\ref{sec:theory}).

\subsubsection*{AI use statement}
Large language model assistants (Claude Fable 5, 5.1 and 5.2) were used throughout this work: to write and debug the experiment, analysis and plotting code, to draft and
edit the text under the authors' direction, and to produce simulated reviews of the draft. The authors designed the experiments, verified
every number against the recorded per-sample outputs, and take full responsibility for the content.

\bibliography{main}
\bibliographystyle{iclr2027_conference}

\appendix
\setlength{\belowcaptionskip}{4pt}
\renewcommand{\floatpagefraction}{0.9}%
\raggedbottom%
\FloatBarrier
\section{The \method{} coefficient step by step}\label{app:algorithm}
Algorithm~\ref{alg:creditfilter} lists the per-batch computation of the coefficient of~\eqref{eq:window}; the lines marked $\star$ are the \method{}-D additions (tracked widths and the class-mass equalization by rescaling, which replaces the floor). Every quantity it uses is already computed by the trainer: the entropies come from the actor's forward pass over the rollouts before the update, the per-token fork
probabilities $C_{i,t}$ come from the same softmax, the advantages and group pass rates come from the reward, and the rank is a sort within each response.

\paragraph{Notation and terms.} Table~\ref{tab:notation} collects the symbols and coined terms of the main text; a few symbols that a reproduced method or an appendix derivation reuses with a local meaning are flagged where they occur.
\begin{table}[htbp]\centering\footnotesize
\caption{Notation and terms.}\label{tab:notation}
\begin{tabular}{@{}L{0.33\linewidth}L{0.63\linewidth}@{}}\toprule
symbol or term & meaning \\ \midrule
$R_i$, $A_i$, $\tilde A_{i,t}=w_{i,t}A_i$ & verifier outcome of response $i$, its group-normalized advantage, and the advantage rescaled by the coefficient $w_{i,t}$ \\
$c(i)$, $s(i)=\mathrm{sign}(A_i)$ & class of a response, its verifier outcome (correct or wrong), which selects the width; sign of its advantage, which selects the floor and the damping \\
$H_{i,t}$, $C_{i,t}=1-\sum_v\pi_v^2$ & entropy and fork probability of token $t$, from the same softmax \\
$q_{i,t}$, $q'$ (midpoint rank) & entropy rank of a token within its response, $1/T_i$ (near $0$) at the highest entropy; the window uses $q=r_t/T_i$, the width fits use the midpoint rank $q'=(r_t-\tfrac12)/T_i$ (Appendix~\ref{app:theory}) \\
$k_{i,t}=\max(2^{-q_{i,t}/\lambda_{c(i)}},\varepsilon_{s(i)})$, $w_c(q)$ & the kernel of a token, with $\lambda$ the width (rank interval over which the weight halves) and $\varepsilon$ the floor; $w_c(q)=\max(2^{-q/\lambda_c},\varepsilon_c)$ is the kernel of a class \\
$M(\lambda,\varepsilon)$ & class mass, the mean of the kernel over the rank (Proposition~\ref{prop:mass}) \\
$g_i=(1-\hat p_i)^\gamma$, $\hat p_i$ & difficulty damping of the responses with a positive advantage; the pass rate of the response's group \\
$m_{\mathrm{gov}}$, $\bar H$, $[H_{lo},H_{hi}]$ & fallback mixing weight ($1$: window on, $0$: uniform credit), batch entropy, and the entropy band that drives it ($\Hlo/\Hhi$ nats) \\
$\nu_t$, $C_t$, $\kappa_t=\nu_t/C_t$ & forking value, fork probability and fork consequence of a position (Section~\ref{sec:theory}) \\
$Q(s,x)$, $V$ & action value and value, the probability of ending correct \\
envelope, $\lambda_{C,c}$; $\lambda_\nu$ & the profile $\mathbb E[C\mid q,c]$ and the width of its exponential fit; the width fitted to the forking-value profile \\
condition K & the mean fork consequence given the rank and $C$ is constant within a class, so the forking spectrum is proportional to the class envelope (Corollary~\ref{cor:window}) \\
$J$ & continuations per branch ($8$ four-branch, $48$ full-support) \\
$m(o)$ & forking spectrum $\mathbb E[\nu_t\mid o]$, the mean-square target for prefix-measurable $o$ with $D$ constant (Appendix~\ref{app:theory}) \\
$\delta_t$; $N(o)$, $D(o)$, $w^*$ & credit increment $Q(s_t,x_t)-V_t$ of the sampled token; $\mathbb E[\delta_tA\mid o]$, $\mathbb E[A^2\mid o]$ and the mean-square coefficient $\max\{0,N/D\}$ (Proposition~\ref{prop:mmse}) \\
value-weighted spectrum, value factor & $\mathbb E[(1-V_1)\,\nu_t/V_t\mid q,c]$ (correct) and $\mathbb E[V_1\,\nu_t/(1-V_t)\mid q,c]$ (wrong), $d$ times the numerator $N(o)$ of the target for the rank and the class under Assumption~\ref{ass:lookahead} (Proposition~\ref{prop:ident}); the weight on $\nu_t$ is the value factor, which the window ignores \\
four-branch, full-support instrument & the sampled token and its three most probable alternatives; the head branches up to mass $0.9$ (at most six) plus two tail draws (Appendix~\ref{app:instrument}) \\
matched-filter gain retained & squared cosine between a weight profile and the measured forking-value profile over the rank strata (Table~\ref{tab:retention}) \\
format tokens & the end-of-text, end-of-turn and thinking-tag tokens; the measured responses have at least 200 other tokens (Appendix~\ref{app:instrument}) \\
\bottomrule\end{tabular}\end{table}

\begin{algorithm}[htbp]
\caption{\method{} coefficient for one training batch, computed from the rollout entropies without gradient (\method{}-D lines marked $\star$)}\label{alg:creditfilter}
\begin{algorithmic}[1]
\Require responses $i=1,\dots,B$ with tokens $t\le T_i$, entropies $H_{i,t}$ and fork probabilities $C_{i,t}=1-\sum_v\pi_{i,t}(v)^2$ from the same softmax, verifier outcomes $R_i$, advantages $A_i$, group pass rates $\hat p_i$; widths $\lambda_\pm$ and floors $\varepsilon_\pm$ ($\star$: initial widths $\lamPos/\lamNeg$, clamps $[\lambda^{\min}_\pm,\lambda^{\max}_\pm]$, floors $0$); fallback band $(H_{\mathrm{lo}},H_{\mathrm{hi}})$; damping exponent $\gamma$
\State $c(i)\gets+$ if $R_i=1$ else $-$;\quad $s(i)\gets\mathrm{sign}(A_i)$;\quad $q_{i,t}\gets\mathrm{rank}_{\downarrow}(H_{i,t}\text{ within }i)/T_i$, ties broken deterministically;\quad $q'_{i,t}\gets q_{i,t}-1/(2T_i)$ \Comment{midpoint rank}
\State $\star$ for each class $c$ with at least 64 responses: average $C$ per response in the ten $q'$-bands of Appendix~\ref{app:track} ($q'\le0.8$), then over responses; if every band mean is present and positive and the least-squares slope $\hat\beta$ of their logarithms on the band centres is negative: $\hat\lambda_{C,c}\gets\mathrm{clip}(-\ln2/\hat\beta;\,\lambda^{\min}_c,\lambda^{\max}_c)$, $\lambda_c\gets(1-\alpha)\lambda_c+\alpha\hat\lambda_{C,c}$ \eqref{eq:track}
\State $k_{i,t}\gets\max(2^{-q_{i,t}/\lambda_{c(i)}},\varepsilon_{s(i)})$ \Comment{kernel of \eqref{eq:window}}
\State $\star$ with the natural mass $M(\lambda,0)=\lambda(1-2^{-1/\lambda})/\ln2$ and $M^{\ast}=\max\{M(\lambda_+,0),M(\lambda_-,0)\}$: $k_{i,t}\gets k_{i,t}\,M^{\ast}/M(\lambda_{c(i)},0)$ for all $i$
\State $\bar H\gets$ token-mean entropy of the retained batch (after dynamic sampling);\quad $m_{\mathrm{gov}}\gets\mathrm{clip}\big((H_{\mathrm{hi}}-\bar H)/(H_{\mathrm{hi}}-H_{\mathrm{lo}}),0,1\big)$
\State $g_i\gets(1-\hat p_i)^{\gamma}$ if $A_i>0$ else $1$
\State $w_{i,t}\gets g_i\,[(1-m_{\mathrm{gov}})+m_{\mathrm{gov}}k_{i,t}]$ \Comment{coefficient of \eqref{eq:window}}
\State maximize the batch token mean of $\min\{\rho_{i,t}w_{i,t}A_i,\ \mathrm{clip}(\rho_{i,t},0.8,1.27)\,w_{i,t}A_i\}$, with the dual clip of Table~\ref{tab:recipe} for $A_i<0$
\end{algorithmic}
\end{algorithm}

\section{Training and evaluation protocol}\label{app:protocol}
This appendix gives the training recipe shared by all runs (Table~\ref{tab:recipe}), the choice of the fallback band, the
evaluation protocol and the statistics.

\begin{table}[htbp]\centering\footnotesize
\caption{Settings shared by all runs; the runs differ only in the entries of Tables~\ref{tab:arms} and~\ref{tab:baselines}.}
\label{tab:recipe}
\setlength{\tabcolsep}{4pt}
\begin{tabular}{@{}lL{11.5cm}@{}}\toprule
Model & Qwen3-8B-Base with its official tokenizer and ChatML chat template \\
Prompts & DAPO-17k~\citep{dapo17k}, boxed-answer variant (17{,}917 problems); prompts of at most 1{,}024 tokens; data seed 201 (300 and 400 in Appendix~\ref{app:seeds}) \\
Batch & 128 prompts $\times$ 16 rollouts per step; mini-batch 32 prompts, i.e.\ four optimizer updates per step \\
Schedule & 400 steps; checkpoint every 25 steps \\
Optimizer & AdamW, learning rate $10^{-6}$ constant (no warm-up), $\beta=(0.9,0.999)$, weight decay $0.01$, gradient clipping at norm $1$, one optimizer pass per mini-batch; no KL loss and no entropy bonus; token-mean loss aggregation \\
Coefficient & the window coefficient is computed under no gradient from the rollout entropies, the verifier outcomes, the sign of each response's advantage and the pass rate of its group; it multiplies the advantage, which is itself a constant of the update \\
Rollouts & vLLM, temperature 1.0, top-$p$ 1.0, no top-$k$; at most 20{,}480 response tokens \\
Reward & training grader (boxed answer, exact match after normalization): $+1$ for a correct and $-1$ for a wrong answer, plus the DAPO soft overlong penalty: with $L_{\max}=20{,}480$ and buffer $L_{\mathrm{c}}=4{,}096$, a response of $\ell$ tokens receives $0$ for $\ell\le L_{\max}-L_{\mathrm c}$, $(L_{\max}-L_{\mathrm c}-\ell)/L_{\mathrm c}$ inside the buffer and $-1$ at the cap (factor 1.0) \\
Sampling & DAPO dynamic sampling: generation chunks of 128 prompts, only prompt groups in which some but not all responses have a positive reward are kept, at most 8 chunks per step \\
Advantage & group mean/standard-deviation normalization (DAPO); PPO ratio $\rho$ clipped to $[1-0.2,\,1+0.27]$ for the \method{} recipe; dual clip: for $A<0$ the per-token objective is $\max\{\min(\rho A,\mathrm{clip}(\rho)A),\,3A\}$ \\
Hardware & three GPU types (H100, H200 and B200); the runs differ only in rollout throughput settings and micro-batching, every other parameter being identical, so we treat them as one training configuration \\
\bottomrule
\end{tabular}
\end{table}

\paragraph{Choice of the fallback band.} The upper bound of the fallback band, $\Hhi$ nats, was set from pilot runs: runs whose batch entropy stayed above $3$ nats for long stretches ended with lower accuracy.

\paragraph{Evaluation protocol.}
Every checkpoint is exported to HF format and sampled with vLLM 0.18.0 (bfloat16, tensor parallel 1, one replica per GPU, seed 0 plus the GPU
index, context 21{,}504 tokens) with $k=32$ samples per problem at temperature 1.0, top-$p$ 0.7 and at most 20{,}480 new tokens, on MATH500
($n=500$), AIME24 ($n=30$) and AIME25 ($n=30$). The last boxed expression is extracted, normalized as in DAPO and graded with Math-Verify~\citep{kydlicek2025mathverify} against the reference (the training reward applies exact
match to the same normalized string). A sample is complete once it has produced a boxed answer; a sample that reaches the 20{,}480-token cap without one is unanswered and counts as wrong, and the tables report the share of such samples. The rule grades the last expression the model boxed, so generation that continues after an answer is not penalized unless it boxes another expression; the overlong penalty of Table~\ref{tab:recipe} shapes training and is not part of the evaluation. All checkpoints are evaluated with this protocol on the same problem sets, and contrasts between two runs are paired by problem.

\paragraph{Statistics.}
The per-problem accuracy of a checkpoint is the mean over its 32 samples; the late average of a run is the mean over checkpoints 350, 375 and
400 of that per-problem accuracy. Intervals are 95\% percentile bootstrap intervals over problems with 3{,}000
resamples; a contrast between two runs resamples problems and takes the mean of the per-problem difference.

\section{Run definitions}\label{app:arms}

All runs share the recipe of Table~\ref{tab:recipe} and differ only in the entries of Table~\ref{tab:arms}; Table~\ref{tab:baselines} gives the settings of the reproduced baselines, Table~\ref{tab:fallback} how often the fallback engaged, and the fidelity notes at the end how each reproduction departs from its publication. The masses $M_+/M_-$ in Table~\ref{tab:arms} follow from the closed form of Proposition~\ref{prop:mass} (Appendix~\ref{app:mass}).

\begin{table}[htbp]\centering\footnotesize
\caption{Run definitions: \method{}, \method{}-D and the uniform-weight recipes (upper block) and the ablations of Table~\ref{tab:ablate} (lower
block); the reproduced baselines are in Table~\ref{tab:baselines}. Kernel: exp $=2^{-q/\lambda_c}$; hard $=\mathbf 1[q\le0.2]$; dyn.\ $=$ tracked
widths, the tracker fitting the fork-probability profile $\mathbb E[C\mid q,c]$ every step; none $=$ uniform weights. Rank: within-response
(resp.), batch-level (batch), random within-response (random) or reversed within-response (reversed) entropy rank. Fallback: band $H_{lo}/H_{hi}$
in nats of the entropy-gated fallback (``off'': $m_{\mathrm{gov}}\equiv1$); Damp.: $\gamma_+/\gamma_-$; Clip: upper clip ratio. $M_+/M_-$:
expected class masses from the closed form on the trainer's rank grid (``$\to$'': natural mass, then rescaled).}
\label{tab:arms}
\setlength{\tabcolsep}{2.5pt}
\begin{tabular}{@{}L{2.75cm}lllllllL{2.35cm}@{}}\toprule
Run & Kernel & Rank & $\lambda_+/\lambda_-$ & $\varepsilon_+/\varepsilon_-$ & Fallback & Damp. & Clip & $M_+/M_-$ \\ \midrule
\method{} & exp & resp. & \lamW/\lamWn & \epsW/0 & \Hlo/\Hhi & 1/0 & 0.27 & 0.196/0.200 \\
\method{}-D$^\dagger$ & exp, dyn. & resp. & \lamPos/\lamNeg & 0/0 & \Hlo/\Hhi & 1/0 & 0.27 & rescaled each step \\
DAPO & none & --- & --- & --- & off & 0/0 & 0.28 & 1/1 \\
GRPO & none & --- & --- & --- & off & 0/0 & 0.20 & 1/1 \\ \midrule
top 20\% window & hard & resp. & --- & 0/0 & \Hlo/\Hhi & 1/0 & 0.27 & 0.200/0.200 \\
wide window & exp & resp. & 0.20/0.28 & \epsW/0 & \Hlo/\Hhi & 1/0 & 0.27 & $0.293\to0.196$ / $0.370\to0.200$ \\
random rank & exp & random & \lamW/\lamWn & \epsW/0 & \Hlo/\Hhi & 1/0 & 0.27 & 0.196/0.200 \\
reversed rank & exp & reversed & \lamW/\lamWn & \epsW/0 & \Hlo/\Hhi & 1/0 & 0.27 & 0.196/0.200 \\
batch-level rank & exp & batch & \lamW/\lamWn & \epsW/0 & \Hlo/\Hhi & 1/0 & 0.27 & 0.196/0.200 under a uniform rank \\
80/20 rule, batch cut & hard & batch & --- & 0/0 & off & 0/0 & 0.28 & 0.200/0.200 in expectation \\
80/20 rule, per-response cut & hard & resp. & --- & 0/0 & off & 0/0 & 0.28 & 0.200/0.200 \\
no floor & exp & resp. & \lamW/\lamWn & 0/0 & \Hlo/\Hhi & 1/0 & 0.27 & 0.144/0.200 \\
rescaled masses & exp & resp. & \lamW/\lamWn & 0/0 & \Hlo/\Hhi & 1/0 & 0.27 & $0.144\to0.200$ / 0.200 \\
no fallback & exp & resp. & \lamW/\lamWn & \epsW/0 & off & 1/0 & 0.27 & 0.196/0.200 \\
no damping & exp & resp. & \lamW/\lamWn & \epsW/0 & \Hlo/\Hhi & 0/0 & 0.27 & 0.196/0.200 \\
damping only (window removed) & none & --- & --- & --- & \Hlo/\Hhi & 1/0 & 0.27 & 1/1 \\
\bottomrule
\end{tabular}

\smallskip
\parbox{\linewidth}{\footnotesize\raggedright $^\dagger$Widths tracked from these initial values: clamps $[0.05,0.30]$/$[0.07,0.40]$, EMA
$\alpha=0.1$, $\ge64$ responses per class (Appendix~\ref{app:track}). Wide window: widths doubled; both class windows rescaled to \method{}'s class masses (shape-only change). Random rank: fresh
permutation of each response's positions every step. Reversed rank: lowest-entropy tokens ranked first. Batch-level rank: per-response mass
varies. Rescaled masses: the fixed-width form of \method{}-D's equalization.
Damping only: uniform token weights; the fallback then has no effect.}
\end{table}

\begin{table}[htbp]\centering\footnotesize
\caption{Reproduced baselines: the 80/20 mask, STARE and SC-GRPO of Table~\ref{tab:main}. Token weight and rank as in
Table~\ref{tab:arms}; host recipe with its upper clip ratio (Clip); none of these runs carries the fallback or the damping (fallback off, damping
$0/0$). Fidelity notes at the end of this appendix.}
\label{tab:baselines}
\setlength{\tabcolsep}{3pt}
\begin{tabular}{@{}lL{2.45cm}lllL{6.3cm}@{}}\toprule
Method & Token weight & Rank & Host & Clip & Settings \\ \midrule
80/20 mask & hard & batch & DAPO & 0.28 & released training script: top-20\% of each micro-batch of 32 responses at every update, by the entropy under the current policy; loss averaged over the kept tokens; 10 warm-up steps, weight decay $0.1$, dual-clip constant $10$; training grader also accepts mathematically equivalent answers \\
STARE & surprisal quantiles & batch & DAPO & 0.28 & O1 variant, $W=1.1$, $M=1.0$, $P=0.10$, $H_{\rm tgt}=0.3$ (STARE's symbols) \\
SC-GRPO & self-conditioned KL & --- & GRPO & 0.20 & $f=D/(D+c)$, $c=\max(P_{75},10^{-4})$, $n_{c,\min}=2$ \\
\bottomrule
\end{tabular}
\end{table}

\begin{table}[htbp]\centering\footnotesize
\caption{Fallback engagement per run, reconstructed from the batch entropy with the band $\Hlo/\Hhi$: mean $m_{\mathrm{gov}}$ and the shares of steps (\%) with the window fully on ($m_{\mathrm{gov}}=1$) and fully replaced by uniform credit ($m_{\mathrm{gov}}=0$). Damping only (uniform weights) is omitted; for the wide-window, batch-level-rank and no-floor runs the available entropy logs cover 352, 371 and 384 of the 400 steps.}
\label{tab:fallback}
\setlength{\tabcolsep}{6pt}
\begin{tabular}{lrrr}\toprule
Run & mean $m_{\mathrm{gov}}$ & fully on (\%) & uniform (\%) \\ \midrule
\FallbackRows
\bottomrule
\end{tabular}
\end{table}

\paragraph{Reproduced baselines: fidelity notes.}
\emph{80/20 mask.} The 80/20 mask of Table~\ref{tab:main} is trained with its authors' released training script on our training set, prompt and schedule (128 prompts per step, 400 steps; Table~\ref{tab:baselines}). The two 80/20-rule rows of Table~\ref{tab:ablate} instead apply the mask's selection rule on the settings of the DAPO row, with the cut taken over the rollout batch or within each response; these two runs normalize the loss over all response tokens instead of over the kept ones (a constant factor $0.2$ that Adam removes; the gradient norm never reached the clipping threshold). \emph{Absolute levels.} Our GRPO run's late MATH500 / AIME24 / AIME25 ($0.833$ / $0.263$ / $0.184$) are close to the $83.0$ / $28.3$ / $20.0$
reported for GRPO on Qwen3-8B-Base by \citet{le2026zvp}, who sample at the same temperature and top-$p$ but score the best checkpoint with 8 samples per problem under an 8{,}192-token cap and clip ratios $0.20/0.28$. The DAPO baseline of \citet{wang2025_8020} on the same
model and prompts, trained with 512 prompts per step for about 40 steps (640 gradient steps) and evaluated at its best checkpoint with 16 samples at temperature $1.0$ (response cap
and top-$p$ not stated), reaches $89.2$ / $33.3$ / $25.4$; \method{}-D reaches $0.887$ / $0.367$ / $0.268$ with 128 prompts per step over 400 steps under
our protocol (top-$p$ $0.7$, 20{,}480-token cap), and our DAPO run at that budget stays at $0.830$ / $0.244$ / $0.177$. The published 80/20 mask
matches its host there ($+0.5$ / $+1.3$ / $+0.8$); trained here with its released script it reaches $0.827$ / $0.302$ / $0.236$.
\emph{STARE.} The O1 variant with the paper's defaults on the DAPO recipe; surprisals are taken from the rollout log-probabilities shared by the
four mini-batches of a step, and the top-$P$ quantile is computed over the batch. Its entropy gate opens only while the batch entropy is below
the target $0.3$ nats, which in our run is the collapsed stretch between steps 10 and 39 (Figure~\ref{fig:train}); elsewhere the run reduces to
its DAPO host. \emph{SC-GRPO.} The self-conditioned teacher (the same
weights, with a correct rollout of the group in the system prompt) and the full-vocabulary forward KL follow the paper; the ``solve-none''
branch does not exist under dynamic sampling (all-wrong groups are removed upstream), the math guide template is ours (the paper's is not
released), responses beyond 8{,}192 (teacher) or 12{,}288 (KL) tokens fall back to unit weight, and the model is the base model rather than the
paper's post-trained one, so the conditioning signal is weaker than in the paper (75th percentile of the KL about $0.007$ in a short test run).

\section{Proofs and computations for Section~\ref{sec:theory}}\label{app:theory}
This appendix proves the propositions of Section~\ref{sec:theory} and the further results stated here (Proposition~\ref{prop:ident}, Corollary~\ref{cor:window},
Proposition~\ref{prop:mass}, Lemma~\ref{lem:adam}, Proposition~\ref{prop:damp}); Appendix~\ref{app:coord} gives the
rationale for the coordinate and the kernel, and Appendix~\ref{app:track} the width tracker and the envelope along training.

\subsection{Setting, sampling law and notation}\label{app:setting}
(S1) A fixed prompt and policy $\pi$; $P$ is the on-policy rollout law; the prefix $s_t$ contains all generated tokens and everything the policy
conditions on. (S2) $s_{t+1}=s_t\circ x_t$; termination is observable from the prefix, $T\le T_{\max}$ almost surely, and $R\in\{0,1\}$ is a deterministic
function of the completed response. Let $\mathcal F_t=\sigma(s_t)$, $V_t=\mathbb E_P[R\mid\mathcal F_t]$, $Q(s_t,a)=\mathbb E_P[R\mid s_t,x_t=a]$,
$\delta_t=V_{t+1}-V_t$, $\nu_t=\mathrm{Var}_{a\sim\pi(\cdot\mid s_t)}Q(s_t,a)$; after termination $V_t=R$ and $\delta_t=\nu_t=0$. Two advantage
conventions: (I) the ideal $A=(R-p)/d$ with $p=V_1$ and fixed $d>0$; (G) the group form $A_i=k(r_i-\bar r_{-i})/d$, $k=(G-1)/G$, $\bar r_{-i}$ the
mean of the other $G-1$ rewards, with $A_i=0$ on homogeneous groups. A \emph{sampling law} over rollout--position pairs is
$\mathbb E_\mu[f]=\mathbb E_P[\sum_{i,t}a_{i,t}f_{i,t}]/Z$ with $a_{i,t}\ge0$, zero after termination, $0<Z<\infty$; it is \emph{predictable} (P) when each
$a_{i,t}$ is $\mathcal F_{i,t}$-measurable (pooled active-token sampling is; drawing a rollout and then a uniform position is predictable only when the length is fixed, since its position weight is $1/T_i$). With
$S_i=r_i-p=\sum_u\delta_{i,u}$: $N(o)=\mathbb E[\delta_tA\mid o]$, $D(o)=\mathbb E[A^2\mid o]$, $m(o)=\mathbb E[\nu_t\mid o]$, the selection correction
$\Delta_1(o)=\mathbb E[\delta_t^2-\nu_t\mid o]$, the cross-position term $\Delta_2(o)=\mathbb E[\delta_t(S_i-\delta_t)\mid o]$ and the peer term
$\Lambda(o)=\mathbb E[\delta_t(\bar r_{-i}-p)\mid o]$. Two rank coordinates are used: the rank $q_t=r_t/T$ (window and masses), with $r_t$ the descending entropy rank within the response and deterministic tie-breaking (Assumption~\ref{ass:lookahead}), and the \emph{midpoint rank} $q'_t=(r_t-\tfrac12)/T$, on which the instrument and the width tracker of \method{}-D bin their profiles; both use all response entropies and are retrospective (prefix-measurable only in degenerate cases such as $T=1$), whereas absolute entropy $H_t$ and a causal rank among positions $\le t$ are prefix-measurable; the outcome class is retrospective as well (prefix-measurable only when $V_1\in\{0,1\}$).

\subsection{Credit decomposition, MMSE and identification (Propositions~\ref{prop:credit}, \ref{prop:mmse} and~\ref{prop:ident})}\label{app:credit}
\paragraph{Decomposition and orthogonality.} $V_t$ is a bounded Doob martingale: $\mathbb E_P[V_{t+1}\mid\mathcal F_t]=V_t$, so
$\mathbb E_P[\delta_t\mid\mathcal F_t]=0$; given $\mathcal F_t$ the next token follows $\pi$ and $V_{t+1}=Q(s_t,x_t)$, so $\mathbb E_P[\delta_t^2\mid\mathcal F_t]=\nu_t$.
For $t<u$, $\delta_t$ is $\mathcal F_u$-measurable and $\mathbb E_P[\delta_t\delta_u]=\mathbb E_P[\delta_t\,\mathbb E_P[\delta_u\mid\mathcal F_u]]=0$. Telescoping,
$\sum_{t\le n}\delta_t=V_{n+1}-V_1$, equal to $R-V_1$ for $n\ge T_{\max}$; squaring and using orthogonality, $\sum_t\mathbb E_P[\nu_t]=\mathrm{Var}_P(R)=p(1-p)$
for the fixed prompt. With almost-surely finite but unbounded $T$, $V_{n+1}=R$ for $n\ge T$, and bounded convergence gives $\mathbb E_P[(V_{n+1}-R)^2]\to0$, so the sums converge
in $L^2$. These statements are about $P$ at fixed indices; retrospective selection can change the expectation measure.

\paragraph{Non-negative coefficient MMSE.} For $w\ge0$ the conditional risk is $\mathbb E[\delta_t^2\mid o]-2wN(o)+w^2D(o)$, minimized on $D>0$ by
$w^*=\max\{0,N/D\}$ (unique a.s.); on $D=0$, $A=0$ a.s.\ conditionally and every finite weight is equivalent. Conditional Cauchy--Schwarz gives
$N^2\le D\,\mathbb E[\delta_t^2\mid o]$, so $w^*A$ has finite second moment. Expanding around $w^*$,
$\mathrm{Risk}(w)-\mathrm{Risk}(w^*)=\mathbb E[D(w-w^*)^2]+2\mathbb E[(-N)w\mathbf 1[N<0]]$, which reduces to $\mathbb E[D(w-w^*)^2]$ when $N\ge0$. This is
coefficient-prediction optimality under the chosen sampling law; the gradient-error bound below relates it to the update.

\paragraph{Clean identification.} Under (I), (P) and $o$ $\mathcal F_t$-measurable: $\mathbb E_P[\delta_t^2-\nu_t\mid\mathcal F_t]=0$; for $u<t$, $\delta_u$ is
$\mathcal F_t$-measurable and $\mathbb E_P[\delta_t\delta_u\mid\mathcal F_t]=0$; for $u>t$,
$\mathbb E_P[\delta_t\delta_u\mid\mathcal F_t]=\mathbb E_P[\mathbb E_P[\delta_t\delta_u\mid\mathcal F_u]\mid\mathcal F_t]=0$. Multiplying each identity by
$a_{i,t}\phi(o_{i,t})$ (an $\mathcal F_{i,t}$-measurable multiplier for bounded $\phi$), summing and dividing by $Z$ transfers them to $\mu$:
$\Delta_1(o)=\Delta_2(o)=0$ and $N(o)=m(o)/d$, hence $w^*(o)=m(o)/(dD(o))$; $m\ge0$, so the positive part is inactive. Proportionality of $w^*$ to
$m$ across observables additionally needs $D$ constant over the compared observables; with a binary reward $D(o)$ depends on $o$ through the outcome class and the pass rate, so the comparison is made within a class (the class masses are set separately, Proposition~\ref{prop:mass}). The within-response rank and the outcome class are retrospective; Proposition~\ref{prop:ident} extends the identification to them under Assumption~\ref{ass:lookahead}.

\paragraph{Finite group and random scale.} Under (G), $\delta_tA_i=(k/d)[\delta_t^2+\delta_t(S_i-\delta_t)-\delta_t(\bar r_{-i}-p)]$, so
$N(o)=(k/d)[m+\Delta_1+\Delta_2-\Lambda](o)$ for any observable and sampling law. If rollouts are i.i.d.\ given the prompt and both $a_{i,t}$ and $o$
depend only on the focal rollout $Z_i$, then $\mathbb E_P[\bar r_{-i}-p\mid Z_i]=0$ and $\Lambda(o)=0$ even for retrospective $o$; including $\mathrm{sign}(A_i)$
or peer information in $o$ can make $\Lambda\ne0$. Under the same hypotheses (i.i.d.\ rollouts, $a_{i,t}$ and $o$ depending only on the focal rollout), for a common positive symmetric scale $d(\mathbf r)$ of the group reward vector (e.g.\ the group standard deviation, with $A_i=0$ on homogeneous groups; deleting such groups instead changes the sampling law) with $G>1$ and $0<p<1$, symmetry and $\sum_iA_i=0$ give $\mathbb E_P[A_i\mid Z_i]=\theta(r_i-p)$ with a prompt-dependent $\theta>0$ (independence is needed for $\theta>0$: if every reward equals one shared Bernoulli variable the group is exchangeable but every advantage vanishes), so $N(o)=\theta[m+\Delta_1+\Delta_2](o)$ for focal-only
observables; $\theta$ must be retained when prompts are pooled.

Assumption~\ref{ass:lookahead} (rank look-ahead) is stated in Section~\ref{sec:theory}; the proposition below is the general form of the identification in Proposition~\ref{prop:mmse}, with the advantage (I) or (G).

\begin{proposition}[identification for the rank and the class, general form]\label{prop:ident}
Let $R$ be the binary verifier outcome, $c$ the class it defines and $A$ the advantage (I) or (G) of $R$, and let the sampling weights be functions of $(s_t,\Gamma_t,R)$ (pooled active tokens and rollout-equal $1/T$ both qualify). Under Assumption~\ref{ass:lookahead}, for a fixed prompt and $o=(q_t,c)$, under (I)
\begin{equation}\label{eq:identgen}
N(o)=\frac1d\,\mathbb E\Big[(1-V_1)\frac{\nu_t}{V_t}\,\Big|\,o\Big]\ \text{on correct responses},\qquad N(o)=\frac1d\,\mathbb E\Big[V_1\frac{\nu_t}{1-V_t}\,\Big|\,o\Big]\ \text{on wrong ones},
\end{equation}
and under (G) the same identities hold with the absolute values of the prompt-level scales $\mathbb E[A\mid\mathrm{prompt},R]$ in place of $(1-V_1)/d$ and $V_1/d$. In both cases $D$ is constant within each class. Hence $w^*$ is proportional to the value-weighted forking spectrum, and to $\mathbb E[\nu_t\mid o]$ when the $\nu$-weighted mean of $1/V_t$ (correct) or $1/(1-V_t)$ (wrong) over the positions at rank $q$ does not depend on $q$.
\end{proposition}
\paragraph{Proof of Proposition~\ref{prop:ident}.} Since $\delta_t$ is $\mathcal F_{t+1}$-measurable,
$\mathbb E_P[\delta_tR\mid\mathcal F_t]=\mathbb E_P[\delta_tV_{t+1}\mid\mathcal F_t]=\mathbb E_P[\delta_t(V_t+\delta_t)\mid\mathcal F_t]=\nu_t$; with $\mathbb E_P[\delta_t\mid\mathcal F_t]=0$ this gives
$\mathbb E_P[\delta_t\mid s_t,R=1]=\nu_t/V_t$ and $\mathbb E_P[\delta_t\mid s_t,R=0]=-\nu_t/(1-V_t)$ on conditioning events of positive probability, and by Assumption~\ref{ass:lookahead} the same values for $\mathbb E_P[\delta_t\mid s_t,\Gamma_t,R]$. On a correct response $A=(1-V_1)/d$ is fixed by the prompt; $o$ and the sampling weights are functions of $(s_t,\Gamma_t,R)$, so the tilt by the weights leaves the conditional law of $\delta_t$ given $(s_t,\Gamma_t,R)$ unchanged, and
$N(o)=\mathbb E_\mu[(1-V_1)\,\mathbb E_P[\delta_t\mid s_t,\Gamma_t,R=1]\mid o]/d=\mathbb E_\mu[(1-V_1)\,\nu_t/V_t\mid o]/d$. With $A=-V_1/d$ the same steps give the wrong-class identity, the negative advantage and the negative conditional increment cancelling. $D$ equals $(1-V_1)^2/d^2$ or $V_1^2/d^2$ on the two classes. For the group advantage the rewards of the other rollouts are independent of the trajectory given the prompt, so the identities hold with the prompt-level scales $\mathbb E[A\mid\mathrm{prompt},R]$ in place of $(1-V_1)/d$ and $-V_1/d$. Strata with $m(o)=0$ have $\nu_t=0$ conditionally, hence $\delta_t=0$ $\mu$-a.s.\ and $N(o)=0$. The overlong penalty of the training reward (Appendix~\ref{app:protocol}), under which $A_i$ depends on the length and $c=\mathrm{sign}(A_i)$ can differ from $R$, lies outside the proposition.

\paragraph{Gradient-error bound.} With $g_t=\nabla\log\pi(x_t\mid s_t)$, $e_t=\delta_t-w_tA$, $\|g_t\|\le B$, positions padded after termination and $L$ the maximum response length:
$\|\sum_te_tg_t\|^2\le(\sum_te_t^2)(\sum_t\|g_t\|^2)\le LB^2\sum_te_t^2$ pathwise, so $\mathbb E\|\sum_te_tg_t\|^2\le LB^2\mathbb E[\sum_te_t^2]$, and by Jensen
the squared norm of the expected discrepancy obeys the same bound. Under differentiability, a valid score identity and policy-independent
reward, $\nabla\mathbb E_P[R]=\mathbb E_P[R\sum_tg_t]=\mathbb E_P[\sum_t\delta_tg_t]$ because $\mathbb E_P[Rg_t]=\mathbb E_P[Q(s_t,x_t)g_t]$ and $\mathbb E_P[V_tg_t]=0$. Under
pooled active-token sampling $\mathbb E_P[\sum_te_t^2]=\mathbb E_P[T]\,\mathrm{Risk}(w)$. The exact discrepancy contains
$\sum_t\mathbb E[e_t^2\|g_t\|^2]+2\sum_{t<u}\mathbb E[e_te_u\langle g_t,g_u\rangle]$; coefficient MMSE ignores score magnitudes, directions and
cross-position cancellation, minimizes only this upper bound, and does not preserve unbiasedness.

\subsection{Factorization and envelope (Proposition~\ref{prop:factor}, Corollary~\ref{cor:window})}\label{app:factor}
\paragraph{Factorization.} Fix a position, write $\pi_k=\pi(k\mid s_t)$ and $Q_k=Q(s_t,k)\in[0,1]$, and let $K,L$ be independent draws from $\pi$.
The variance of a random variable is half the mean squared difference of two independent copies, so
$\nu=\mathrm{Var}\,Q_K=\tfrac12\mathbb E[(Q_K-Q_L)^2]=\tfrac12\sum_{k,l}\pi_k\pi_l(Q_k-Q_l)^2$. Diagonal terms vanish, hence
$\nu=\mathbb P(K\ne L)\cdot\tfrac12\mathbb E[(Q_K-Q_L)^2\mid K\ne L]=C\kappa$ with $C=\mathbb P(K\ne L)=1-\sum_k\pi_k^2$, and $\kappa:=0$ when $C=0$ (a
deterministic policy has $\nu=0$). Since $(Q_k-Q_l)^2\in[0,1]$, $\kappa\in[0,\tfrac12]$. In words, $C$ is the probability that two independent
draws of the policy differ (the fork probability) and $\kappa$ is half the mean squared change in expected correctness caused by such a fork
(the fork consequence).

\paragraph{Envelope chain.} $\nu=C\kappa\le C/2$. From $\sum_k\pi_k^2\ge\pi_{\max}^2$,
$C\le1-\pi_{\max}^2=(1-\pi_{\max})(1+\pi_{\max})\le2(1-\pi_{\max})$. From $\sum_k\pi_k^2\le\pi_{\max}\sum_k\pi_k=\pi_{\max}$, $1-\pi_{\max}\le C$. For the
entropy, $-\ln$ is convex, so Jensen gives $H=\mathbb E[-\ln\pi_K]\ge-\ln\mathbb E[\pi_K]=-\ln\sum_k\pi_k^2=-\ln(1-C)\ge C$, using $-\ln(1-x)\ge x$ on
$[0,1)$; the intermediate quantity $-\ln\sum_k\pi_k^2$ is the collision entropy. Equality $1-\pi_{\max}=C$ holds iff $\pi$ is uniform on its
support, and $C=H$ only for a deterministic law. Two further bounds: $\nu\le V_t(1-V_t)$ because $Q_K\in[0,1]$ has mean $V_t$ (Bhatia--Davis),
and $\sum_t\mathbb E[\nu_t]=p(1-p)$ (Proposition~\ref{prop:credit}), so the total forking value of a prompt vanishes as its pass rate approaches
$0$ or $1$. All quantities in each identity refer to one and the same action law; a restricted-law $\nu$ cannot replace the full-policy $\nu$
in the martingale budget.

\begin{corollary}[condition K: the envelope window]\label{cor:window}
If within a class $\mathbb E[\kappa\mid q,c,C]=\bar\kappa_c>0$ depends neither on the rank nor on $C$ (condition K), then
$\mathbb E[\nu\mid q,c]=\bar\kappa_c\,\mathbb E[C\mid q,c]$. If, within a class, $\mathbb E[\kappa_t/V_t\mid q,+,C]$ (correct) and $\mathbb E[\kappa_t/(1-V_t)\mid q,-,C]$ (wrong) depend neither on the rank nor on $C$ (condition K for the value-weighted consequence), then under Assumption~\ref{ass:lookahead}
\begin{equation}
w^*(q,c)\ \propto\ \mathbb E[C\mid q,c]:
\end{equation}
the window is the class envelope profile.
\end{corollary}
\paragraph{Condition K and the envelope window.} Within a class,
$\mathbb E[\nu\mid q,c]=\mathbb E[C\kappa\mid q,c]=\mathbb E[C\,\mathbb E[\kappa\mid q,c,C]\mid q,c]$. Under condition K, $\mathbb E[\kappa\mid q,c,C]=\bar\kappa_c$ (it need
only be imposed where $C>0$), so $\mathbb E[\nu\mid q,c]=\bar\kappa_c\,\mathbb E[C\mid q,c]$; a decay scale is identified only when $\bar\kappa_c>0$, the envelope profile is not identically zero and at least two ranks occur (constant action values give $\kappa=m=w^*=0$ despite $C>0$). Under the clean identification of Appendix~\ref{app:credit} (prefix-measurable
observable, predictable sampling, constant $D$), $w^*=m/(dD)\propto\mathbb E[C\mid q,c]$; for the rank and the outcome class, Proposition~\ref{prop:ident} gives the value-weighted numerators $\mathbb E[(1-V_1)\,C_t\kappa_t/V_t\mid q,+]/d$ and $\mathbb E[V_1\,C_t\kappa_t/(1-V_t)\mid q,-]/d$ with $D$ constant within a class, and conditioning on $C$ as above turns them into multiples of $\mathbb E[C\mid q,c]$ if condition K also holds for the value-weighted consequence; condition K alone does not imply this. In
both cases the shape within a class is the envelope profile; the criterion fixes the class amplitudes (Section~\ref{sec:theory}) and the class mass rule sets the implementation amplitudes equal. Wherever
$\mathbb E[C\mid q,c]>0$, the proportionality holds exactly when the $C$-weighted band mean $\bar\kappa_c(q)=\mathbb E[C\kappa\mid q,c]/\mathbb E[C\mid q,c]$ is rank-independent, and condition K is sufficient for it; this ratio is
what the instrument estimates as $\mathbb E[\hat\nu\mid q,c]/\mathbb E[C\mid q,c]$, and it differs from the unweighted mean of $\kappa$ when $\kappa$ and $C$
are correlated within a band. Condition K is substantive: for two independent fair tokens with reward equal to the second token, both
positions have $C=\tfrac12$ and $H=\ln2$ while $\nu_1=0$ and $\nu_2=\tfrac14$, so identical envelopes can carry different forking value and an envelope
alone cannot identify it. If $C_{0,c}>0$, at least two ranks occur and $\mathbb E[C\mid q,c]=C_{0,c}2^{-q/\lambda_{C,c}}$ exactly, the window is the exponential family with $\lambda_c=\lambda_{C,c}$ and no floor; with a fitted exponential, $\lambda_C$ is the decay scale of the ten-band fit (Appendix~\ref{app:track}). Section~\ref{sec:spectrum} and Appendix~\ref{app:instrument} assess condition K; Appendix~\ref{app:omitted} measures how much the value weighting changes the spectrum.

\subsection{Window mass and the positive-class floor (Proposition~\ref{prop:mass})}\label{app:mass}
\begin{proposition}[window mass and the positive-class floor]\label{prop:mass}
Conditional on the length $T$, the class and the window parameters, a response's mean kernel value does not depend on its entropy values and
lies within $[w(0)-w(1)]/T$ of a continuum mass $M(\lambda,\varepsilon)$ with a closed form (below); for $\lambda_->\lambda_+$ the
equal-mass condition of~\eqref{eq:mass} has a unique solution on the active-floor interval.
\end{proposition}
\paragraph{Per-response mass.} With ranks $q_t=r_t/T$ the ranks of a response of length $T$ are a permutation of $\{j/T:j=1,\dots,T\}$, so
conditional on $T$, the class and the window parameters the response's mean kernel value
$\tfrac1T\sum_tw(q_t)=M_T(\lambda,\varepsilon)=\tfrac1T\sum_{j=1}^T\max\{2^{-j/(\lambda T)},\varepsilon\}$ is independent of its entropy values. $M_T$ depends on
$T$ and converges to the continuum mass $M(\lambda,\varepsilon)=\int_0^1\max\{2^{-q/\lambda},\varepsilon\}\,dq$ with $0\le M-M_T\le[w(0)-w(1)]/T$, because
the right-endpoint sum of a non-increasing integrand lies below its integral by at most one cell of height $w(0)-w(1)$. The ratio of
normalized class mean weights therefore approaches $M(\lambda_+,\varepsilon_+)/M(\lambda_-,\varepsilon_-)$ when both classes are present with
sufficiently long responses; total class masses also scale with the class token counts and are not equalized. A batch-wide rank does not have the per-response property (the batch-level rank ablation of Table~\ref{tab:ablate}).

\paragraph{Closed form.} For $0<\lambda<\infty$, $0\le\varepsilon\le1$ and uniform $q$ on $[0,1]$, let $b=2^{-1/\lambda}$ and $q^*=-\lambda\log_2\varepsilon$ (the point where the exponential meets the floor; $q^*=+\infty$ when $\varepsilon=0$). If $\varepsilon\le b$ the floor is inactive and $M=\int_0^12^{-q/\lambda}dq=\tfrac{\lambda}{\ln2}(1-b)$;
if $b\le\varepsilon\le1$, $M=\int_0^{q^*}2^{-q/\lambda}dq+\varepsilon(1-q^*)=\tfrac{\lambda}{\ln2}(1-\varepsilon)+\varepsilon(1+\lambda\log_2\varepsilon)$, using $2^{-q^*/\lambda}=\varepsilon$;
the two branches agree at $\varepsilon=b$, and the floored expression is not valid outside $[b,1]$. Values: $M(0.10,0.1)=0.1966$ ($q^*=0.332$),
$M(0.14,0)=0.2005$, $M(0.10,0)=0.1441$, so the floor $\epsW$ of the \method{} configuration adds $0.0525$ of continuum mass, all of it on $q\ge0.33$.

\paragraph{Positive-class floor.} For fixed $\lambda$, $\varepsilon\mapsto M(\lambda,\varepsilon)$ is continuous, constant on $[0,b]$ and strictly increasing on $[b,1]$ ($\partial M/\partial\varepsilon=1+\lambda\log_2\varepsilon=1-q^*$, zero at $\varepsilon=b$ and positive for $\varepsilon>b$) up to $M(\lambda,1)=1$, and $\lambda\mapsto M(\lambda,0)$ is strictly increasing. Hence
for $\lambda_->\lambda_+$ the equation $M(\lambda_+,\varepsilon_+)=M(\lambda_-,0)$ has exactly one solution on the active-floor interval $[b_+,1)$ (for
$\lambda_-=\lambda_+$ every $\varepsilon\in[0,b_+]$ solves it, so uniqueness comes only from selecting the active-floor interval), and by implicit
differentiation the solution increases when $\lambda_+$ narrows or $\lambda_-$ widens. Bisection gives $\varepsilon_+=0.106$ for $(0.10,0.14)$ (target
mass $0.2005$); the floor $0.10$ of the \method{} widths equalizes only approximately (continuum
mass ratio $0.1966/0.2005=0.98$). Within the clamps of \method{}-D a floor equalizes the classes only by flattening the window: $\lambda_+=0.05$ with $\lambda_-=0.40$ needs mass $0.475$, and even a floor of $0.3$ gives only $M(0.05,0.3)=0.324$; \method{}-D therefore drops the floor and equalizes the classes by rescaling (Section~\ref{sec:dynamic}).

\subsection{Why the within-response rank and the exponential kernel}\label{app:coord}
The window's coordinate and kernel follow from three requirements on a per-token coefficient, none of which uses the measurement.

\paragraph{A rank weight redistributes credit within a response.} By Proposition~\ref{prop:credit} the expected forking value of a prompt's rollouts, summed over positions, is $V(1-V)$. The identity fixes
this mean over rollouts, not the value of a single response, and a rank weight redistributes it within each response. A
coefficient that is a function of the entropy values gives two responses of the same length and outcome different totals: for $w_t=C_t$ the total is the
length times the response's mean fork probability, which moves by an order of magnitude along training, so the size of a response's update would follow its uncertainty. A function of the
within-response ranks leaves the total a function of the length alone (Proposition~\ref{prop:mass}), so the window's shape is one object across responses, training regimes and models.

\paragraph{The rank is invariant and bounded.} The rank of a token within its response is the same under any monotone transform of the
uncertainty measure (the entropy, its exponential the perplexity, any power of either), whereas a value-based coefficient commits to one scale
of the softmax. A kernel on the rank lies in $[0,1]$ by construction, which keeps the coefficient sign-preserving, the premise of
Proposition~\ref{prop:mmse}, and bounded by one, so no token receives more than the shared advantage (the class-mass rescaling of \method{}-D raises the head weights of one class above one); a $z$-score of the entropy within the response is signed, and a
ratio to the response mean is unbounded.

\paragraph{The exponential is the memoryless kernel.} Among kernels on the rank, the exponential $w(q)=2^{-q/\lambda}$ is the one whose weight
ratio between two positions depends only on their rank difference, $w(q+\delta)/w(q)=2^{-\delta/\lambda}$ (Cauchy's equation for a monotone
kernel): it prefers no scale along the rank, and one number, the width, sets the concentration. This is what makes the width a per-model
calibration that a tracker can re-measure in the loop and that Table~\ref{tab:models} compares across models, and it gives the mass its closed
form (Appendix~\ref{app:mass}). The measured envelope has this form ($R^2$ $0.99$ and $0.95$ over the ten bands, the same form on eight models),
so the fitted width is the calibration.

\paragraph{Derived, measured, chosen and approximated parts.} Table~\ref{tab:parts} lists the parts of \method{} by their source.

\begin{table}[H]\centering\footnotesize
\caption{The parts of \method{} by their source, and where each is stated, ablated or tested.}
\label{tab:parts}
\setlength{\tabcolsep}{4pt}
\begin{tabular}{@{}L{1.9cm}L{7.2cm}L{4.0cm}@{}}\toprule
Source & Part & Where \\ \midrule
Derived & The target of a sign-preserving token weight is the value-weighted forking spectrum; it reduces to the forking spectrum, and to the fork-probability envelope, under the stated conditions & Propositions~\ref{prop:mmse} and~\ref{prop:factor}, Corollary~\ref{cor:window} \\
Measured & The fork-probability envelope over the within-response rank: exponential, with widths $\lamPos$ and $\lamNeg$ on the base model (the runs use $\lamW$ and $\lamWn$, the rounded widths of a subset of the responses; Section~\ref{sec:window}); the fork consequence over the rank & Section~\ref{sec:spectrum}, Tables~\ref{tab:spectrum} and~\ref{tab:fine} \\
Chosen & The within-response rank and the exponential form of the window, from three requirements that do not use the measurement; the class-mass floor, the damping and the fallback & This subsection; ablated in Table~\ref{tab:ablate} \\
Approximated & Assumption~\ref{ass:lookahead}; the fork consequence taken as flat over the rank (condition K); the value factor left out of the window & Appendices~\ref{app:atwo}, \ref{app:fine} and~\ref{app:omitted} \\
\bottomrule
\end{tabular}
\end{table}

\subsection{Lemma on Adam and gradient scale}\label{app:adam}
\begin{lemma}[Adam and gradient scale]\label{lem:adam}
With gradient clipping inactive and zero initial moments, on coordinates with nonzero second moment (a zero coordinate steps by zero under the convention $0/0:=0$), multiplying the entire gradient history by a constant $k>0$ leaves the bias-corrected Adam update \citep{kingma2015adam}
unchanged at $\epsilon_{\mathrm{Adam}}=0$ (and replaces $\epsilon_{\mathrm{Adam}}$ by $\epsilon_{\mathrm{Adam}}/k$ otherwise). For a constant signed coordinate
gradient multiplied by $k\ge1$ after $t_0$ baseline updates, at $\epsilon_{\mathrm{Adam}}=0$ the update $n$ steps later relative to the unchanged
reference is $F_n=[1+(k-1)a_1]/\sqrt{1+(k^2-1)a_2}\le k$, $a_j=(1-\beta_j^n)/(1-\beta_j^{t_0+n})$.
\end{lemma}
Adam keeps its moment estimates (a local use of $M$ and $V$) $M_t=\beta_1M_{t-1}+(1-\beta_1)g_t$, $V_t=\beta_2V_{t-1}+(1-\beta_2)g_t^2$ and steps by $\alpha\hat M_t/(\sqrt{\hat V_t}+\epsilon_{\mathrm{Adam}})$
with the usual bias corrections; replacing $g_t$ by $kg_t$ throughout gives $kM_t$ and $k^2V_t$, hence the first claim (decoupled weight decay
is unaffected). For the step change, $M_{t_0+n}=\bar g(1-\beta_1^{t_0+n})[1+(k-1)a_1]$ and $V_{t_0+n}=\bar g^2(1-\beta_2^{t_0+n})[1+(k^2-1)a_2]$,
giving $F_n$; with $r=\epsilon_{\mathrm{Adam}}/|\bar g|$ the general ratio is $[1+(k-1)a_1](1+r)/[\sqrt{1+(k^2-1)a_2}+r]$, still at most $k$. For
$k=4$, $\beta_1=0.9$, $\beta_2=0.999$, $t_0=100$ and $r=0$: $F_n=1.89,1.59,1.34,1.12$ at $n=10,50,100,300$, peak $1.93$ near $n=15$; with
equilibrated moments the peak is $3.23$. \emph{Consequence for the design.} A common amplitude fixed for the whole run is therefore immaterial under these assumptions but not outside them (a factor introduced mid-run, a non-negligible $\epsilon_{\mathrm{Adam}}$, active clipping), whereas changing the ratio of the class mean weights can change the aggregate gradient direction and Adam's moments. \method{} equalizes the class mean weights with the floor, which places the equalizing mass on the tail. The rescaled-mass ablation of Table~\ref{tab:ablate} places it on the head, and the no-floor ablation leaves the masses unequal.

\subsection{Tracking the envelope}\label{app:track}
\paragraph{Estimator and update.} At every step, for each class $c$ with at least $64$ responses in the batch, the tokens of each response are binned by their midpoint rank $q'$ into the ten bands with edges $(0,.025,.05,.075,.1,.15,.2,.3,.4,.6,.8)$, the band mean of the fork probability $C=1-\sum_v\pi_v^2$ (per token, from the same softmax pass as the entropies) is taken per response, the response means are averaged over the responses that have the band (rollout-equal), and $\hat\lambda_{C,c,t}=\ln2/(-\text{slope})$ from an ordinary least-squares fit of the log band means on the band centers; the width is updated only when every band mean is present and positive and the fitted slope is finite and negative; otherwise, including for a flat profile, the previous width is retained. The tracked width is the exponential moving average $\lambda_{c,t}=(1-\alpha)\lambda_{c,t-1}+\alpha\,\hat\lambda_{C,c,t}$ with $\alpha=0.1$, the target and the average both clamped to $[0.05,0.30]$ (correct) and $[0.07,0.40]$ (wrong) in the \method{}-D run, whose initial widths are $\lamPos/\lamNeg$; the state is persisted so that a restart resumes the average.

\paragraph{Envelope along training.} Table~\ref{tab:track} applies this estimator to training-length rollouts of the \method{} checkpoints every 50 steps. At step 350, in the aftermath of
the entropy excursion, 28\% of the rollouts hit the cap and are excluded, and the width of the remaining, shorter wrong responses ($0.08$) is
not comparable with the other rows. \method{}-D's tracker, which follows the envelope of its own batches, has widened the wrong-class
width to its clamp of $0.40$ by step 290 and ends training at $0.28$--$0.37$ over its last 75 updates; its correct-class width is $0.16$ at
step 290 and $0.13$--$0.18$ at the end. The widths that fail when fixed from the first step, $0.20/0.28$ (Table~\ref{tab:ablate}), are close to those
the tracker reaches after step 290. The fixed \method{} window stays close to the moving envelope: its squared cosine with the exponential of the
tracked width is $0.84$--$0.95$ for the correct class and $0.83$--$1.00$ for the wrong class across the checkpoints of this table, against
$0.65$--$0.91$ and $0.86$--$1.00$ for the wide window and $0.66$--$0.81$ and $0.55$--$0.81$ for the top 20\% window.
\begin{table}[t]\centering\footnotesize
\caption{Envelope widths along the \method{} run on training-length rollouts of its checkpoints (224 questions $\times$ 4 rollouts, 20{,}480-token
cap; the tracker's estimator with 95\% rollout-bootstrap intervals over 300 replicates). $n$: rollouts per class among those that finished
under the cap; capped: share of the 896 rollouts cut at the cap and excluded ($^\dagger$: more than a fifth).
Base-policy widths for reference: $\lamPos$ (correct) and $\lamNeg$ (wrong), Table~\ref{tab:spectrum}.}\label{tab:track}
\setlength{\tabcolsep}{5pt}
\begin{tabular}{@{}lrlrlr@{}}\toprule
 & \multicolumn{2}{c}{correct} & \multicolumn{2}{c}{wrong} & \\
\cmidrule(lr){2-3}\cmidrule(lr){4-5}
step & $n$ & $\lambda_C$ & $n$ & $\lambda_C$ & capped (\%) \\ \midrule
\TrackRows
\bottomrule
\end{tabular}
\end{table}

\subsection{Damping multiplier (Proposition~\ref{prop:damp})}\label{app:damp}
\begin{proposition}[damping multiplier]\label{prop:damp}
For binary $R$ with pass probability $p\in(0,1)$, $A=(R-p)/\sqrt{p(1-p)}$ and an unclipped full-trajectory score $S$, under a valid score identity with a policy-independent reward and with $p$, the normalization and the damping coefficients held fixed, scaling correct responses by $(1-p)^{\gamma}$ gives $\mathbb E[g(R)AS]=[p+(1-p)^{\gamma+1}]\,\nabla p/\sqrt{p(1-p)}$.
\end{proposition}
\paragraph{Damping.} With true pass probability $0<p<1$, $A=(R-p)/\sigma$, $\sigma=\sqrt{p(1-p)}$, the unclipped full-trajectory score $S$, a valid score identity with a policy-independent reward, and $p$, $\sigma$ and the damping coefficients treated as fixed: $g(1)=(1-p)^{\gamma}$, $g(0)=1$; $\mathbb E[RS]=\nabla p$ and, since $\mathbb E[S]=0$, $\mathbb E[(1-R)S]=-\nabla p$; on $R=1$, $g(R)A=(1-p)^{\gamma+1}/\sigma$ and on $R=0$, $g(R)A=-p/\sigma$, so $\mathbb E[g(R)AS]=\sigma^{-1}\{(1-p)^{\gamma+1}\nabla p+p\nabla p\}=[p+(1-p)^{\gamma+1}]\nabla p/\sigma$. For $\gamma=1$ the multiplier $1-p+p^2=\tfrac34+(p-\tfrac12)^2$ lies in $[\tfrac34,1)$; the population ascent direction is preserved. The identity does not cover estimated
group pass rates, random normalization, windowing, clipping or Adam.

\section{The forking-value instrument}\label{app:instrument}
This appendix describes the two instruments of Section~\ref{sec:spectrum}: the four-branch instrument, whose measurement on Qwen3-8B-Base
is summarized in Table~\ref{tab:spectrum} and drawn band by band in Figure~\ref{fig:factor}, and the full-support instrument of
Appendix~\ref{app:fine}.

\begin{table}[htbp]\centering\footnotesize
\caption{The forking spectrum of Qwen3-8B-Base by response class (Figure~\ref{fig:main}a and Figure~\ref{fig:factor}; the numbers behind Section~\ref{sec:spectrum}): widths of the fork-probability envelope and of
the forking value; the fork consequence pooled over bands 2--10, its relative slope (count-weighted least-squares slope of $\hat\kappa$ over the band centres of bands 2--10, divided by $\bar\kappa$) and the top-band gain; 95\% rollout-bootstrap intervals.
$\hat\kappa$ and $\hat\nu$ in units of $10^{-3}$.}
\label{tab:spectrum}
\begin{tabular}{lcc}\toprule
 & correct & wrong \\ \midrule
responses / positions & 294 / 7056 & 672 / 16128 \\
envelope width $\lambda_C$ & \lamPos{} [0.099, 0.123] & \lamNeg{} [0.146, 0.165] \\
forking-value width $\lambda_\nu$ & 0.125 [0.063, 0.207] & 0.146 [0.086, 0.332] \\
top-band forking value $\mathbb E[\hat\nu\mid q\le0.025]$ & 3.93 [2.64, 5.24] & 1.07 [0.55, 1.52] \\
consequence $\bar\kappa$ (bands 2--10) & 2.12 [1.21, 3.13] & 0.70 [0.43, 0.98] \\
relative slope of $\hat\kappa$ (bands 2--10) & $-2.1$ [$-5.2$, 1.1] & 0.3 [$-3.3$, 4.6] \\
top-band gain $\hat\kappa_1/\bar\kappa$ & 2.9 [1.7, 5.6] & 2.4 [1.3, 3.9] \\
\bottomrule
\end{tabular}
\end{table}

\subsection{Estimator}
\paragraph{Setting.} At a position $t$ with prefix $s_t$ and next-token law $\pi$, the branch set is the sampled token together with the
three most probable alternatives under $\pi$ (four branches). With $\mathrm{cover}=\sum_k\pi(x_k)$ the weights are $\omega_k=\pi(x_k)/\mathrm{cover}$. Each
branch is continued $J$ times, independently, with the training sampler (temperature $1$, top-$p$ $1.0$, a budget of the instrument's 4608-token context
less the prefix), and every continuation is graded $g_{k,j}\in\{0,1\}$ by the training grader (strict boxed-answer match against the
reference); a length-truncated continuation is graded $0$. Let $Q_k=\mathbb E[g_{k,j}]$ be the action value of branch $k$ under that sampler,
$S_k=\sum_jg_{k,j}$, $\hat Q_k=S_k/J$ and $U_k=S_k(S_k-1)/(J(J-1))$. The estimate is
\begin{equation}\label{eq:nuhat}
\hat\nu_t=\sum_k \omega_k(1-\omega_k)\,U_k-\sum_{k\ne l}\omega_k\omega_l\,\hat Q_k\hat Q_l,
\end{equation}
signed, and exactly zero when all $4J$ continuations agree.

\paragraph{Unbiasedness.} $U_k$ is the fraction of ordered pairs of distinct continuations of branch $k$ that both succeed, a U-statistic of
order two, so $\mathbb E[U_k]=Q_k^2$; independence across branches gives $\mathbb E[\hat Q_k\hat Q_l]=Q_kQ_l$ for $k\ne l$. Hence the estimate
$\hat\nu_t$ of~\eqref{eq:nuhat} has expectation
\[
\begin{aligned}
\mathbb E\Big[\sum_k\omega_k(1-\omega_k)U_k-\sum_{k\ne l}\omega_k\omega_l\hat Q_k\hat Q_l\Big]
&=\sum_k\omega_kQ_k^2-\sum_k\omega_k^2Q_k^2-\sum_{k\ne l}\omega_k\omega_lQ_kQ_l\\
&=\sum_k\omega_kQ_k^2-\Big(\sum_k\omega_kQ_k\Big)^2,
\end{aligned}
\]
the variance of $Q$ under $\omega$ on the branch support: the quantity $\nu_t$ of Section~\ref{sec:theory} restricted to the retained support, with
$C_t=1-\sum_k\omega_k^2$ the matching fork probability (the identity $\nu=C\kappa$ of Proposition~\ref{prop:factor} holds on any finite support).
The instrument therefore measures $\nu$, $C$ and $\kappa$ under the renormalized four-branch law; the omitted mass $1-\mathrm{cover}$ can carry forking value that the retained support does not show, and the full-support instrument of Appendix~\ref{app:fine} includes it. The band estimator $\hat\kappa=\mathbb E[\hat\nu\mid q]/\mathbb E[C\mid q]$ inherits the sign of $\hat\nu$ and can be negative (Table~\ref{tab:bands}), although $\kappa=\nu/C\in[0,\tfrac12]$.

\begin{figure}[tbp]\centering
\includegraphics[width=\linewidth]{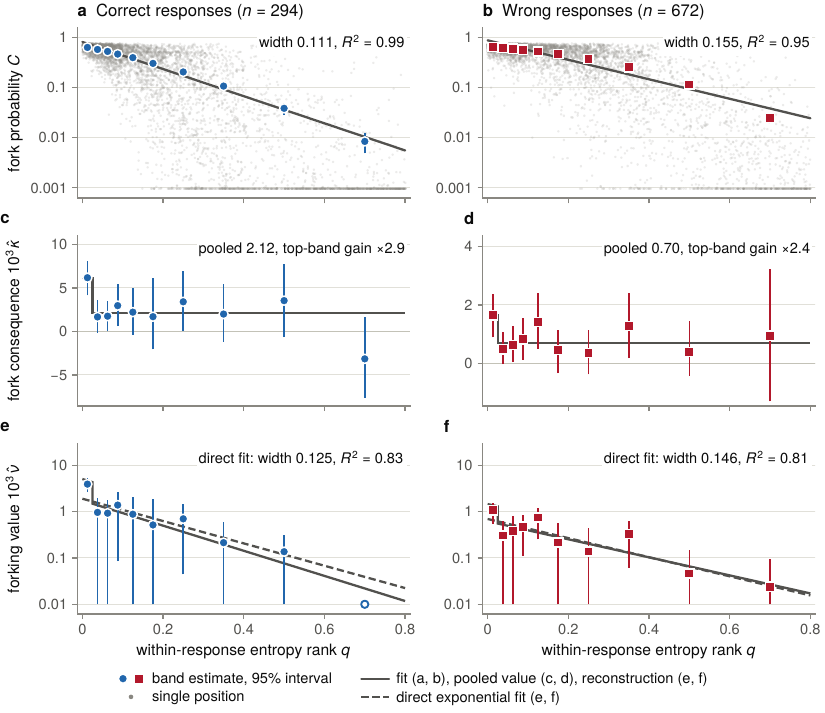}
\caption{The factorization $\nu=C\kappa$ on Qwen3-8B-Base. Left: correct responses; right: wrong responses; vertical lines: 95\% rollout-bootstrap intervals. (a,\,b)~Fork probability $C$: gray dots, 6000 positions per class, with values below $10^{-3}$ drawn at $10^{-3}$; markers, band means; line, the exponential fit, with $\lambda_C=\lamPos$, $R^2=0.99$ for correct and $\lambda_C=\lamNeg$, $R^2=0.95$ for wrong responses. (c,\,d)~Fork consequence $\hat\kappa(q)=\mathbb E[\hat\nu\mid q]/\mathbb E[C\mid q]$ per band; line, the pooled value $\bar\kappa$ over bands 2--10 with the top-band gain on band 1; the values are in Table~\ref{tab:spectrum}. (e,\,f)~Forking value $\mathbb E[\hat\nu\mid q]$ per band, hollow at the floor for a band mean $\le0$; solid line, the reconstruction $\bar\kappa\,\hat C(q)$ with the top-band gain; dashed line, the direct exponential fit, with $\lambda_\nu=0.125$, $R^2=0.83$ for correct and $\lambda_\nu=0.146$, $R^2=0.81$ for wrong responses.}\label{fig:factor}
\end{figure}

\subsection{Cover and truncation}
Cover, the probability mass retained by the four branches, has mean $0.902$ (correct) and $0.795$ (wrong) and rises with the rank, from $0.69/0.48$ in the top band to at least $0.96$ beyond $q=0.3$: the low-cover positions are those of the head, where the policy spreads mass beyond four tokens, and the top-band gain of Table~\ref{tab:spectrum} is a statement about $0.69/0.48$ of the mass at those positions. Summed over positions, $\hat\nu$ is $2.4$ (correct) and $3.0$ (wrong) times the budget $V(1-V)$ of Proposition~\ref{prop:credit}; the window uses only the shape of each class profile, which a class-wide factor leaves unchanged. Of the $4J=32$ continuations per position,
$2.1\%/3.7\%$ are length-truncated and graded $0$, $35\%/48\%$ of positions have at least one truncated continuation, and $89.7\%/73.3\%$ of
continuations end with a boxed answer.

\paragraph{Continuation horizon.}
The base-policy continuations run inside the 4{,}608-token context and a continuation that reaches its end is graded $0$, whereas training
generates under a 20{,}480-token budget. Table~\ref{tab:horizon} repeats the spectrum statistics under two alternative treatments of the stored
records: only the positions none of whose 32 continuations was truncated ($65\%$ of the correct-class and $52\%$ of the wrong-class
positions), and every truncated continuation graded $1$ instead of $0$. The envelope width does not move (it depends on the softmax only); the forking-value width stays inside its interval, and its interval contains the envelope width; the pooled consequence moves within a factor of
$1.5$ with overlapping intervals; the top-band gain stays above one in every treatment; the head-to-tail ratio of $\hat\kappa$ resolves no
trend in any; and the class ratio $\bar\kappa_+/\bar\kappa_-$ ranges from $3.0$ to $4.8$.

\begin{table}[htbp]\centering\scriptsize
\caption{Continuation-horizon check on the base policy: the spectrum statistics under three treatments of the length-truncated continuations
(positions kept; envelope and forking-value widths over the ten bands; fork consequence pooled over bands 2--10, in units of $10^{-3}$; top-band gain
$\hat\kappa_1/\bar\kappa$; head-to-tail ratio of $\hat\kappa$, $q\le0.2$ against $q>0.2$ within bands 2--10; 95\% rollout bootstrap, replicated independently of Table~\ref{tab:spectrum}).}
\label{tab:horizon}
\setlength{\tabcolsep}{3pt}
\begin{tabular}{@{}lL{1.9cm}rlllll@{}}\toprule
class & treatment & positions & $\lambda_C$ & $\lambda_\nu$ & $\bar\kappa$ & top-band gain & head-to-tail \\ \midrule
\HorizonRows
\bottomrule
\end{tabular}
\end{table}

\subsection{Protocol}
\paragraph{Corpus.} 288 training prompts with 8 responses each at temperature $1$, top-$p$ $1.0$ ($13.7\%$ of responses correct); 640 responses were
selected with at least 200 non-format tokens and prompt plus response within 4160 tokens (294 correct, 346 wrong). The wrong class is the union of the wrong responses from mixed-outcome prompts of this draw and of a second draw of the same prompts, 672
responses. Format tokens are excluded from the rank; positions carry the midpoint rank $q'=(r-\tfrac12)/T$ of Appendix~\ref{app:setting}, with $r$ the 1-based descending
entropy rank, $T$ the number of ranked tokens and ties broken by position.
\paragraph{Positions.} Per response, $3,3,3,2,2,2,3,2,2,2$ positions are drawn uniformly at random from the ten bands
$(0,.025],\ (.025,.05],\ \dots,\ (.6,.8]$, 24 in total, so every response has the same design in every band, and a band mean
$\sum_rS_{r,b}/\sum_rN_{r,b}$ over responses $r$ ($S_{r,b}$: the sum of the estimates at the positions of response $r$ in band $b$; $N_{r,b}$: their number) weights responses equally (the per-band counts of Table~\ref{tab:bands} differ from this design because positions are re-binned by the midpoint rank $q'$, which shifts some positions across band edges); the tail $q>0.8$ is not sampled. Table~\ref{tab:bands} lists the
ten-band measurement for both classes.

\begin{table}[htbp]\centering\footnotesize
\caption{Ten-band measurement on Qwen3-8B-Base (top: 294 correct responses, 7056 positions; bottom: 672 wrong responses, 16128 positions).
$n$: positions per band. Band means with 95\% rollout-bootstrap intervals of the forking value $\hat\nu$, the fork probability $C$, the entropy $H$
(nats) and the fork consequence $\hat\kappa=\mathbb E[\hat\nu\mid q]/\mathbb E[C\mid q]$; $\hat\nu$ and $\hat\kappa$ in units of $10^{-3}$.}\label{tab:bands}
\setlength{\tabcolsep}{4.5pt}
\begin{tabular}{@{}lrllll@{}}\toprule
band $q$ & $n$ & $\mathbb E[\hat\nu\mid q]$ & $\mathbb E[C\mid q]$ & $\mathbb E[H\mid q]$ & $\hat\kappa(q)$ \\ \midrule
\multicolumn{6}{l}{\emph{correct responses}} \\
0--0.025 & 904 & 3.93 [2.64, 5.24] & 0.637 [0.630, 0.643] & 2.49 [2.35, 2.62] & 6.17 [4.14, 8.12] \\
0.025--0.05 & 886 & 0.96 [$-$0.08, 1.96] & 0.579 [0.570, 0.589] & 1.81 [1.70, 1.92] & 1.66 [$-$0.18, 3.58] \\
0.05--0.075 & 875 & 0.93 [0.00, 1.81] & 0.530 [0.517, 0.542] & 1.51 [1.40, 1.62] & 1.75 [$-$0.01, 3.49] \\
0.075--0.1 & 580 & 1.39 [0.09, 2.68] & 0.469 [0.452, 0.488] & 1.28 [1.18, 1.38] & 2.96 [0.60, 5.39] \\
0.1--0.15 & 582 & 0.88 [$-$0.03, 2.05] & 0.398 [0.376, 0.419] & 1.05 [0.95, 1.15] & 2.21 [$-$0.36, 4.98] \\
0.15--0.2 & 587 & 0.52 [$-$0.58, 1.85] & 0.306 [0.285, 0.328] & 0.76 [0.69, 0.83] & 1.69 [$-$2.06, 6.08] \\
0.2--0.3 & 880 & 0.70 [0.04, 1.50] & 0.206 [0.185, 0.231] & 0.50 [0.44, 0.58] & 3.40 [0.09, 6.94] \\
0.3--0.4 & 586 & 0.21 [$-$0.16, 0.61] & 0.107 [0.091, 0.125] & 0.28 [0.23, 0.34] & 1.99 [$-$1.18, 5.44] \\
0.4--0.6 & 589 & 0.14 [$-$0.04, 0.31] & 0.039 [0.028, 0.050] & 0.11 [0.08, 0.14] & 3.53 [$-$0.65, 7.69] \\
0.6--0.8 & 587 & $-$0.03 [$-$0.08, 0.01] & 0.008 [0.005, 0.012] & 0.02 [0.02, 0.03] & $-$3.15 [$-$7.65, 1.68] \\
\midrule
\multicolumn{6}{l}{\emph{wrong responses}} \\
0--0.025 & 2071 & 1.07 [0.55, 1.52] & 0.647 [0.644, 0.651] & 3.65 [3.55, 3.77] & 1.65 [0.90, 2.37] \\
0.025--0.05 & 2011 & 0.31 [$-$0.02, 0.62] & 0.628 [0.623, 0.633] & 2.81 [2.71, 2.91] & 0.49 [$-$0.02, 1.08] \\
0.05--0.075 & 2005 & 0.38 [$-$0.01, 0.79] & 0.602 [0.594, 0.608] & 2.41 [2.32, 2.50] & 0.63 [0.05, 1.27] \\
0.075--0.1 & 1323 & 0.47 [0.11, 0.86] & 0.572 [0.564, 0.581] & 2.12 [2.03, 2.22] & 0.83 [0.14, 1.55] \\
0.1--0.15 & 1347 & 0.75 [0.25, 1.22] & 0.527 [0.517, 0.537] & 1.77 [1.69, 1.85] & 1.42 [0.51, 2.40] \\
0.15--0.2 & 1335 & 0.22 [$-$0.12, 0.58] & 0.468 [0.457, 0.481] & 1.45 [1.38, 1.53] & 0.46 [$-$0.33, 1.16] \\
0.2--0.3 & 2015 & 0.14 [$-$0.13, 0.44] & 0.377 [0.364, 0.391] & 1.06 [1.00, 1.13] & 0.37 [$-$0.37, 1.16] \\
0.3--0.4 & 1342 & 0.33 [0.06, 0.63] & 0.255 [0.241, 0.270] & 0.70 [0.64, 0.76] & 1.29 [0.19, 2.40] \\
0.4--0.6 & 1337 & 0.05 [$-$0.04, 0.14] & 0.115 [0.105, 0.128] & 0.34 [0.30, 0.39] & 0.40 [$-$0.41, 1.45] \\
0.6--0.8 & 1342 & 0.02 [$-$0.03, 0.09] & 0.025 [0.020, 0.031] & 0.09 [0.07, 0.12] & 0.95 [$-$1.29, 3.22] \\
\bottomrule\end{tabular}
\end{table}

\paragraph{Widths.} $\lambda_X=\ln2/(-\text{slope})$ of an ordinary least-squares fit of $\ln\mathbb E[X\mid q]$ on the band centers over the bands
with positive mean, ten bands unless stated (for $X=C$ every band mean is positive; for $X=\hat\nu$ a negative tail band is dropped, which can lengthen $\lambda_\nu$, whereas $\lambda_C$, the quantity the window uses, is unaffected).
The log band means of $C$ are concave: the local half-width falls from $0.18$--$0.20$ over the first bands to $0.09$ in the last for correct responses and from $0.58$ to $0.09$ for wrong ones, and a fit restricted to $q\le0.2$ gives widths $0.15$/$0.34$ against the global widths of Table~\ref{tab:spectrum}; the single exponential is a global summary, narrower than the head and wider than the tail.
\begin{table}[htbp]\centering\footnotesize
\caption{Forking spectrum on further models with the same instrument (120 responses $\times$ 24 positions, 60 for Qwen3-32B, four branches, $J=8$, top-$p$ $1.0$;
classes pooled): per model, the number of responses, the ten-band forking-value width $\lambda_\nu$ with its interval, the envelope width $\lambda_C$
at the same positions, and the coarse-group means of $\hat\nu$ ($G_1$: $q\le0.1$, $G_2$: $0.1<q\le0.4$, $G_3$: $q>0.4$). With 60--120 responses the
forking-value width intervals are wide, whereas $\lambda_C$ is resolved on every model. A width point estimate is undefined when the ten-band fit does not decay; its interval is then over the replicates whose fit does.}\label{tab:models}
\begin{tabular}{@{}lrlll@{}}\toprule
model & $n$ & $\lambda_\nu$ [95\% CI] & $\lambda_C$ & $\mathbb E[\hat\nu\mid G]$ ($10^{-3}$): $G_1/G_2/G_3$ \\ \midrule
Qwen3-1.7B & 120 & 0.119 [0.047, 1.057] & 0.071 & 0.65 / 0.44 / $-$0.03 \\
Qwen3-4B & 120 & -- [0.062, $>2$] & 0.086 & 0.37 / 0.26 / $-$0.02 \\
Qwen3-14B-Base & 120 & 0.255 [0.072, 0.913] & 0.108 & 0.79 / 0.04 / 0.16 \\
Qwen3-32B & 60 & 0.307 [0.059, $>2$] & 0.091 & 0.25 / 0.17 / 0.03 \\
Qwen3-8B (post-trained) & 120 & 0.218 [0.043, 1.499] & 0.084 & 0.31 / 0.34 / $-$0.01 \\
Qwen2.5-7B-Instruct & 120 & 0.130 [0.048, 1.225] & 0.081 & 0.41 / 0.12 / $-$0.02 \\
Llama-3.1-8B-Instruct & 120 & 0.306 [0.094, 1.592] & 0.267 & 0.68 / 0.25 / 0.20 \\
\bottomrule\end{tabular}
\end{table}

\paragraph{Summary.} The width of an exponential fitted to the forking value itself lies within $20\%$ of the envelope's, though its intervals span a factor of three to four (Table~\ref{tab:spectrum}). Under the full-support instrument, the top $20\%$ of the rank captures $62\%$ (correct) and $82\%$ (wrong) of the forking value (Table~\ref{tab:fine}).

\subsection{A full-support instrument at higher resolution}\label{app:fine}
\paragraph{Design.} The full-support instrument is run once on the base policy. Responses: 16
rollouts of 512 training prompts at the training sampler with a 5{,}800-token cap; from the mixed-outcome prompts, 400 responses per class (at
most two per prompt and class, at least 200 ranked tokens; 296/297 prompts). Positions: one uniform draw from each of 16 rank strata per
response (edges $0$, $0.0025$, $0.005$, $0.01$, $0.015$, $0.025$, $0.0375$, $0.05$, $0.075$, $0.1$, $0.15$, $0.2$, $0.3$, $0.4$, $0.6$, $0.8$, $1$)
plus one draw from each of the top global-entropy strata, with recorded inclusion probabilities; every bin is a Horvitz--Thompson mean, so the
estimates are consistent in any coordinate. Branch law: the most probable tokens in descending order until their mass reaches $0.9$, with at least two and at most six
branches (so the head covers less than $0.9$ at the highest-entropy positions, column cover of Table~\ref{tab:fine}), plus two independent draws from the remaining mass $r$; with $D_{ab}=U_a+U_b-2\hat Q_a\hat Q_b$ the estimate
$\hat\nu=\sum_{a<b}p_ap_bD_{ab}+r\sum_kp_k(D_{k,y_1}+D_{k,y_2})/2+\tfrac12r^2D_{y_1,y_2}[y_1\ne y_2]$ is unbiased for the full-policy $\nu$
(checked on $4\times10^4$ random laws), and the head-restricted pair $(\hat\nu_{\mathrm{head}},C_{\mathrm{head}})$ is kept for comparison with
the four-branch instrument. Each branch is continued $J=48$ times under the training sampler with a budget of $5{,}800$ tokens minus the
position index; $C$, $H$ and $\pi$ come from a forward pass over every token. Responses are weighted so that a class-$c$ response of a prompt
stands for all eligible class-$c$ rollouts of that prompt, the population a training batch sees; intervals are 95\% cluster bootstraps over
prompts (2{,}000 replicates). The run took $3.58$ million continuations and $2.97$ billion generated tokens on 16 GPUs in $6.2$ hours;
$1.7\%/2.7\%$ of the continuations hit the budget.

\begin{table}[htbp]\centering\scriptsize
\caption{The full-support instrument on the base policy, training-population weights: per class and rank stratum, sampled positions, cover
of the head branches, fork probability $C$ over all positions of the stratum, forking value $\hat\nu$ and consequence $\hat\kappa$ under the full law
(in units of $10^{-3}$, 95\% prompt bootstrap) and the consequence under the head-restricted law.}
\label{tab:fine}
\setlength{\tabcolsep}{3pt}
\begin{tabular}{llrrrlll}\toprule
class & rank $q$ & $n$ & cover & $C$ & $\hat\nu$ [95\%] & $\hat\kappa$ [95\%] & $\hat\kappa_{\mathrm{head}}$ \\ \midrule
\FineRows
\bottomrule
\end{tabular}
\end{table}

\paragraph{Where the consequence concentrates (Table~\ref{tab:fine}).} The fork consequence is concentrated in the top $0.25\%$ of the rank, about two tokens per response. Pooled over ranks $0.5$--$2.5\%$ it is $3.29$ $[2.17, 4.81]$ against $1.56$ $[1.12, 2.08]$ over ranks $2.5$--$20\%$, a
ratio of $2.10$ $[1.29, 3.45]$; for wrong responses $0.73$ $[0.44, 1.13]$ against $0.31$ $[0.18, 0.47]$, ratio $2.36$ $[1.27, 4.62]$. Beyond
$q=0.2$ the fork probability is below $0.17$ for correct and $0.39$ for wrong responses, and the estimates of $\kappa=\nu/C$ widen: $3.79$ $[1.56, 7.33]$ for correct responses (the $2.5$--$20\%$ value is $0.41$ $[0.19, 1.05]$ of it; the estimate is carried by the strata $0.3$--$0.6$ whose $C$ is $0.03$--$0.08$) and $0.13$ $[0.01, 0.31]$ for wrong ones (the $2.5$--$20\%$ value is $2.4$ $[0.9, 15]$ times it). Under question-balanced weights the head ratios are $1.53$ $[0.99, 2.36]$ and $2.26$ $[1.20, 4.40]$. In
the top stratum the full-law consequence lies below the head-law one ($9.45$ against $11.45$ for correct and $1.95$ against $2.23$ for wrong
responses), so the probability mass beyond the head dilutes the fork there; below it the two laws are close. Aggregated to the ten bands of the
four-branch instrument, the top band gives $4.05$ $[3.04, 5.31]$ against $6.17$ $[4.14, 8.12]$ there, overlapping. On this corpus the envelope
widths over the ten bands are $0.092$ (correct) and $0.138$ (wrong) and the forking-value widths $0.080$ and $0.113$: the forking value decays
somewhat faster than the fork probability, and the envelope width is a property of the corpus and sampler on which it is measured (the
four-branch corpus gives $\lamPos$ $[0.099, 0.123]$), which is why the width is calibrated per model and the adaptive variant re-measures it.

\begin{table}[htbp]\centering\footnotesize
\caption{Matched-filter gain retained by weight profiles against the forking-value profile of the full-support instrument (16 strata,
width-weighted; the exact profile retains $1$). Brackets: 95\% question-cluster bootstrap intervals (2{,}000 replicates).}
\label{tab:retention}
\begin{tabular}{lcc}\toprule
weight profile & correct & wrong \\ \midrule
\RetentionRows
\bottomrule
\end{tabular}
\end{table}

\paragraph{What the envelope window retains (Table~\ref{tab:retention}).} With the forking-value profile $m(q)=\mathbb E[\hat\nu\mid q]$ of the 16
strata as the target and the strata widths as position weights, a weight profile $w$ retains the fraction
$\langle w,m\rangle^2/(\|w\|^2\|m\|^2)$ of the matched-filter gain of $m$ itself (the mean-square criterion of Proposition~\ref{prop:mmse}
with $D$ constant, ignoring the value factor). An exponential refitted to the forking value gains little over the window, so the width is not the limiting factor, whereas a threefold gain on the top $0.25\%$ gains more; the piecewise profile $C\kappa$ with the measured consequence per region is fitted on the same estimates it is scored against, so its value is an in-sample reference. Question-bootstrap intervals (Table~\ref{tab:retention}) are wide, and resampling adds noise to the target, so they lie below the point values; the paired differences are better resolved: the window retains $0.21$ $[0.13, 0.25]$ (correct) and $0.13$ $[0.02, 0.23]$ (wrong) more than the top 20\% window, $0.40$ $[0.20, 0.50]$ and $0.48$ $[0.33, 0.51]$ more than uniform weights, and $0.04$ $[-0.05, 0.11]$ and $0.16$ $[0.09, 0.18]$ more than the wide window. The top-rank gain is not part of \method{}: the top stratum is where the four-branch cover is lowest and it holds two tokens per response.

\subsection{A test of Assumption~\ref{ass:lookahead}}\label{app:atwo}
Assumption~\ref{ass:lookahead} states that, given the prefix and the outcome, the rank carries no further information about the increment $\delta_t$. We test it on the instrument of Appendix~\ref{app:fine} (800 base-policy responses, $13{,}129$ positions), at the $10{,}925$ positions whose sampled token is one of the branched tokens, where the realized increment is measured. Within each outcome class we regress the realized increment on indicators of rank bands, controlling for the token entropy. Under the assumption the conditional mean of the increment is $\nu_t/V_t$ in correct and $-\nu_t/(1-V_t)$ in wrong responses (Appendix~\ref{app:credit}). Least squares is linear in the dependent variable, so each coefficient splits exactly into the coefficient of this implied increment and a remainder that the assumption sets to zero. Table~\ref{tab:atwo} gives both, with 95\% intervals from a bootstrap over prompts.

At equal entropy the top-ranked tokens carry larger increments, and most of this association is what the assumption itself implies. Over the top $2.5\%$ of the rank the remainder is within sampling error in both classes. By finer band it shows only in the top $0.25\%$ of correct responses, with the sign of the advantage: the realized increment there is about $1.5$ times the implied one. Counting the remainder, the top $2.5\%$ of the rank holds $28\%$ of the correct-class target instead of $25\%$ (the strata's mean increments, realized against implied, summed over the rank with the strata's widths as weights), so the approximation puts slightly too little weight at the head, where the window already puts its largest weight.

\begin{table}[H]\centering\footnotesize
\caption{Test of Assumption~\ref{ass:lookahead} on the base policy: the coefficient of each rank band, the part the assumption implies and the remainder (95\% intervals in brackets). $^{\ast}$: one indicator for the top $2.5\%$ of the rank; other rows: four bands, against the remaining $70\%$. $n$: positions in the band.}
\label{tab:atwo}
\setlength{\tabcolsep}{4pt}
\resizebox{\linewidth}{!}{\begin{tabular}{@{}lrlll@{}}\toprule
Rank band & $n$ & Coefficient & Implied part & Remainder \\ \midrule
\multicolumn{5}{@{}l}{\emph{Correct responses}} \\
top 2.5\%$^{\ast}$ & 1{,}512 & $+$0.0091 [0.0041, 0.0140] & $+$0.0061 [0.0037, 0.0088] & $+$0.0030 [$-$0.0014, 0.0072] \\
top 0.25\% & 275 & $+$0.0398 [0.0259, 0.0534] & $+$0.0267 [0.0191, 0.0350] & $+$0.0131 [0.0029, 0.0231] \\
0.25--2.5\% & 1{,}237 & $+$0.0058 [$-$0.0004, 0.0122] & $+$0.0035 [0.0001, 0.0073] & $+$0.0022 [$-$0.0033, 0.0075] \\
2.5--10\% & 1{,}426 & $+$0.0006 [$-$0.0037, 0.0054] & $+$0.0000 [$-$0.0025, 0.0027] & $+$0.0006 [$-$0.0028, 0.0040] \\
10--30\% & 1{,}162 & $-$0.0005 [$-$0.0029, 0.0021] & $-$0.0004 [$-$0.0016, 0.0008] & $-$0.0001 [$-$0.0023, 0.0022] \\
\midrule
\multicolumn{5}{@{}l}{\emph{Wrong responses}} \\
top 2.5\%$^{\ast}$ & 1{,}293 & $-$0.0035 [$-$0.0071, 0.0000] & $-$0.0019 [$-$0.0030, $-$0.0009] & $-$0.0016 [$-$0.0049, 0.0015] \\
top 0.25\% & 253 & $-$0.0045 [$-$0.0141, 0.0050] & $-$0.0048 [$-$0.0083, $-$0.0012] & $+$0.0003 [$-$0.0082, 0.0094] \\
0.25--2.5\% & 1{,}040 & $-$0.0051 [$-$0.0100, $-$0.0008] & $-$0.0022 [$-$0.0038, $-$0.0006] & $-$0.0029 [$-$0.0073, 0.0011] \\
2.5--10\% & 1{,}276 & $-$0.0024 [$-$0.0057, 0.0010] & $-$0.0006 [$-$0.0019, 0.0006] & $-$0.0018 [$-$0.0046, 0.0010] \\
10--30\% & 1{,}093 & $+$0.0005 [$-$0.0016, 0.0025] & $-$0.0005 [$-$0.0011, 0.0001] & $+$0.0010 [$-$0.0010, 0.0029] \\
\bottomrule
\end{tabular}}
\end{table}

\subsection{The omitted value factor}\label{app:omitted}
The target of Proposition~\ref{prop:mmse} weights the forking value by $1/V_t$ in correct and by $1/(1-V_t)$ in wrong responses, a factor the window leaves out. On the same instrument both spectra are stratum means over the 16 rank strata of Table~\ref{tab:fine} with the weights of Table~\ref{tab:retention}; in the value-weighted one each position's forking value is divided by $V_t$ or $1-V_t$, estimated from the position's continuations and floored at $0.05$. Weighting each stratum by its width over the rank, the two spectra have a squared cosine of $0.969$ $[0.898, 0.985]$ in correct and $0.981$ $[0.933, 0.992]$ in wrong responses (95\% intervals from a bootstrap over prompts): the factor barely changes the shape of the target over the rank.

\subsection{Bootstrap}
All intervals of the four-branch instrument (Tables~\ref{tab:spectrum}--\ref{tab:models}) are 95\% percentile intervals of a bootstrap over responses (the 24 positions of a response are resampled together): 400
replicates for band means, 600 for the $\kappa$ statistics and 300 for widths, where replicates whose fit does not decay are excluded (none for the base spectra and envelopes).

\section{Additional training results}\label{app:extra}
\begin{figure}[htbp]\centering
\includegraphics[width=\linewidth]{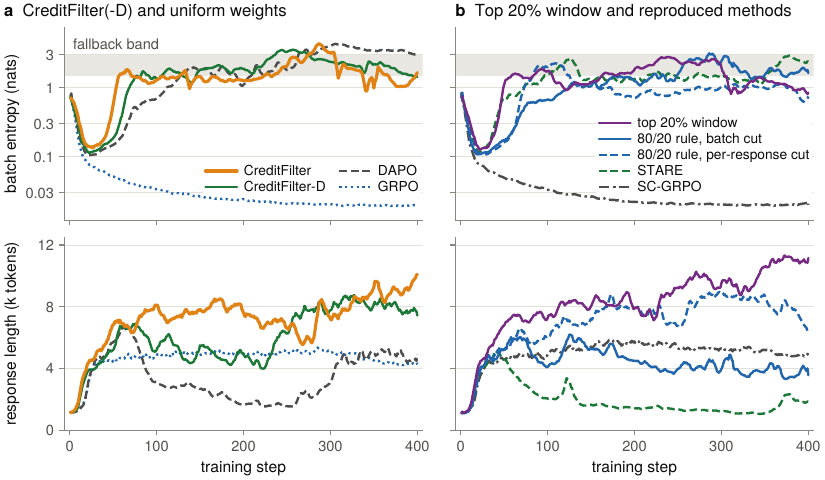}
\caption{Training dynamics of \method{} against the baselines. Top: batch entropy as a five-step moving average on a log axis, with the fallback band shaded. Bottom: mean
response length as a five-step moving average. (a) \method{} and \method{}-D against the uniform-weight recipes DAPO and GRPO; (b) the top 20\% window, the 80/20 rule on the settings of the DAPO row, with the cut taken over the rollout batch or within each response, and the
reproduced methods STARE and SC-GRPO.}
\label{fig:train}
\end{figure}

This appendix gives the training curves of nine runs (Figure~\ref{fig:train}), the lengths and unanswered shares of the step-400 checkpoints (Table~\ref{tab:trunc}), the training cost with the comparisons at matched budgets (Appendix~\ref{app:cost}) and the runs with further data seeds (Appendix~\ref{app:seeds}).

\paragraph{Length and unanswered samples.}
The share of samples that reach the cap (Table~\ref{tab:trunc}) is larger than the unanswered share: at the late checkpoints \method{} reaches the cap in $3.8$--$20.6\%$ of its MATH500 samples
and leaves $0.3$--$1.8\%$ unanswered, and its checkpoint at step 350, taken shortly after the entropy excursion of Figure~\ref{fig:train}, has the
largest cap share and $1.8$ / $5.4$ / $3.0\%$ unanswered.
The base model already reaches the cap after having boxed an answer in $3.8$--$5.0\%$ of its samples, and the trained runs inherit the habit: at step 400, $3.3$--$3.6\%$ of \method{}'s samples and $7.5$--$11.6\%$ of
DAPO's reach the cap after their answer, while GRPO, whose entropy has collapsed (Figure~\ref{fig:train}), stops at the answer ($0.0$--$0.3\%$).
Generated length therefore mixes reasoning length with continuation after the answer.

\begin{table}[htbp]\centering\footnotesize
\caption{Step-400 checkpoints under the mean@32 protocol: mean generated tokens and share of samples (\%) that reach the 20{,}480-token cap without a boxed answer (unans.) or after having boxed one (at cap), per set
(MATH500 $n=500$, AIME24 $n=30$, AIME25 $n=30$; 32 samples per problem). The base model is given at step 0.}
\label{tab:trunc}
\setlength{\tabcolsep}{3pt}
\begin{tabular}{@{}L{4.0cm}rrrrrrrrr@{}}\toprule
 & \multicolumn{3}{c}{MATH500} & \multicolumn{3}{c}{AIME24} & \multicolumn{3}{c}{AIME25} \\
\cmidrule(lr){2-4}\cmidrule(lr){5-7}\cmidrule(l){8-10}
Run & tokens & unans. & at cap & tokens & unans. & at cap & tokens & unans. & at cap \\ \midrule
\TruncRows
\bottomrule
\end{tabular}
\end{table}

\subsection{Training cost and matched budgets}\label{app:cost}
\paragraph{Training cost.} Every step trains on the same number of prompts and responses, in the same number of updates, so a run's training tokens follow the length of its responses. Table~\ref{tab:cost} gives each run's tokens through the optimizer over the 400 steps. \method{}, GRPO and DAPO were trained on the same 16 H100 GPUs, where these steps took about $145$, $80$ and $70$ hours: \method{} trains on $2.1$ times DAPO's tokens in $2.1$ times its time, at about the same time per token. At \method{}'s time per token, the $6.88$ billion tokens of the top 20\% window correspond to about $165$ hours. The tokens alone do not carry the accuracy: the random-rank run of Table~\ref{tab:ablate}, on \method{}'s settings, trains on about $5.0$ billion tokens and scores $0.849$ / $0.255$ / $0.217$.

\begin{table}[htbp]\centering\footnotesize
\caption{Training tokens over 400 steps: tokens through the optimizer, in billions.}
\label{tab:cost}
\begin{tabular}{lr@{\hskip 3em}lr}\toprule
Run & tokens ($10^9$) & Run & tokens ($10^9$) \\ \midrule
\method{} & 6.04 & 80/20 mask & 5.21 \\
\method{}-D & 5.06 & SC-GRPO & 4.19 \\
GRPO & 3.90 & STARE & 1.72 \\
DAPO & 2.87 & top 20\% window & 6.88 \\
\bottomrule
\end{tabular}
\end{table}

\paragraph{Equal decoding budgets.} Table~\ref{tab:budget} samples the responses of the late checkpoints anew, cuts each response at a token budget and grades it as in the main protocol; its full-length row differs from Table~\ref{tab:main} by at most one point. \method{} leads from a budget of 8k tokens upward and trails at 4k, where about $70\%$ of its AIME responses have not yet written an answer. At equal tokens spent the lead remains: a majority vote over three DAPO samples (the most frequent answer among three of the 32 samples of Table~\ref{tab:main}'s evaluation, ties broken at random) spends $7.3$k / $13.7$k / $9.6$k tokens per problem, more than one \method{} response ($6.4$k / $9.1$k / $8.4$k), and reaches $0.858$ / $0.264$ / $0.190$; a vote over eight samples reaches $0.874$ / $0.303$ / $0.212$. A single \method{} sample scores $0.876$ / $0.338$ / $0.277$ (Table~\ref{tab:main}).

\begin{table}[htbp]\centering\footnotesize
\caption{Accuracy at equal decoding budgets (mean@32, late average): each response cut at the budget. MATH500 / AIME24 / AIME25.}
\label{tab:budget}
\begin{tabular}{lccc}\toprule
Budget & \method{} & DAPO & GRPO \\ \midrule
4k & 0.779 / 0.168 / 0.176 & 0.827 / 0.242 / 0.179 & 0.806 / 0.208 / 0.168 \\
8k & 0.867 / 0.286 / 0.258 & 0.828 / 0.243 / 0.179 & 0.832 / 0.253 / 0.193 \\
12k & 0.875 / 0.326 / 0.274 & 0.828 / 0.243 / 0.179 & 0.832 / 0.255 / 0.194 \\
20{,}480 & 0.876 / 0.334 / 0.277 & 0.828 / 0.243 / 0.179 & 0.832 / 0.255 / 0.194 \\
\bottomrule
\end{tabular}
\end{table}

\paragraph{Equal training tokens.} We continued the DAPO run from its step-400 checkpoint with unchanged settings (Table~\ref{tab:dapoext}). More training does not raise DAPO's accuracy: up to step 600, at $1.56$ times its training tokens at step 400, it stays near its step-400 level. At equal training tokens \method{} leads: at $4.2$ billion tokens its step-300 checkpoint scores $0.850$ / $0.299$ / $0.250$ against $0.829$ / $0.256$ / $0.204$ for DAPO's step 575.

\begin{table}[H]\centering\footnotesize
\caption{The DAPO run continued from step 400, and \method{}'s checkpoint at equal training tokens: cumulative training tokens and mean@32.}
\label{tab:dapoext}
\begin{tabular}{lrrccc}\toprule
Run & Step & tokens ($10^9$) & MATH500 & AIME24 & AIME25 \\ \midrule
DAPO & 400 & 2.87 & 0.837 & 0.259 & 0.195 \\
 & 500 & 3.65 & 0.838 & 0.266 & 0.188 \\
 & 525 & 3.82 & 0.840 & 0.259 & 0.201 \\
 & 550 & 3.99 & 0.837 & 0.275 & 0.210 \\
 & 575 & 4.18 & 0.829 & 0.256 & 0.204 \\
 & 600 & 4.48 & 0.832 & 0.256 & 0.206 \\
\midrule
\method{} & 300 & 4.25 & 0.850 & 0.299 & 0.250 \\
\bottomrule
\end{tabular}
\end{table}
\paragraph{Exponential window against the top 20\% window.} Table~\ref{tab:expvstop} compares the two windows under the same damping, fallback, clip and class masses, over the three data seeds of Appendix~\ref{app:seeds}. In every seed and on every set the exponential window generates fewer tokens at evaluation ($17$--$28\%$ fewer on average) and leaves fewer responses unanswered at the cap, and in every seed it trains on fewer tokens ($15\%$ fewer on average). On the AIME sets, counting only the responses that end within 8k tokens, the exponential window gives more correct answers in every seed ($0.277$ / $0.223$ against $0.229$ / $0.195$ on AIME24 / AIME25, means over the seeds); within 12k tokens the two windows are level ($0.305$ / $0.241$ against $0.293$ / $0.240$).

\begin{table}[H]\centering\footnotesize
\caption{Exponential window (\method{}) against the within-response top 20\% window, mean $\pm$ standard deviation over the three data seeds of Table~\ref{tab:seeds}: late averages under the mean@32 protocol, and training tokens over 400 steps.}
\label{tab:expvstop}
\begin{tabular}{llccc}\toprule
 & Window & MATH500 & AIME24 & AIME25 \\ \midrule
accuracy & exponential & $0.860\pm0.025$ & $0.326\pm0.042$ & $0.258\pm0.035$ \\
 & top 20\% & $0.817\pm0.084$ & $0.347\pm0.050$ & $0.261\pm0.026$ \\
mean generated tokens & exponential & $6{,}541\pm281$ & $8{,}611\pm979$ & $7{,}491\pm1{,}201$ \\
 & top 20\% & $7{,}894\pm677$ & $11{,}730\pm841$ & $10{,}428\pm951$ \\
unanswered at the cap (\%) & exponential & $1.2\pm0.6$ & $4.6\pm0.8$ & $1.7\pm0.4$ \\
 & top 20\% & $6.4\pm7.0$ & $12.1\pm1.6$ & $9.9\pm1.9$ \\ \midrule
training tokens ($10^9$) & exponential & \multicolumn{3}{c}{$5.57\pm0.58$} \\
 & top 20\% & \multicolumn{3}{c}{$6.58\pm0.34$} \\
\bottomrule
\end{tabular}
\end{table}

\paragraph{Length up to the answer.} The generated lengths of Table~\ref{tab:trunc} include text written after the answer. Table~\ref{tab:anslen} counts each response only up to where it first states its final answer: \method{}'s reasoning up to the answer is $2.8$--$4.3$ times DAPO's and $1.5$--$2.0$ times GRPO's.

\begin{table}[H]\centering\footnotesize
\caption{Mean number of tokens up to the first statement of the final answer, over the responses that state one (late average, mean@32 protocol).}
\label{tab:anslen}
\begin{tabular}{lrrr}\toprule
Run & MATH500 & AIME24 & AIME25 \\ \midrule
\method{} & 2{,}556 & 7{,}028 & 6{,}607 \\
DAPO & 905 & 1{,}969 & 1{,}549 \\
GRPO & 1{,}696 & 4{,}080 & 3{,}300 \\
\bottomrule
\end{tabular}
\end{table}

\FloatBarrier
\subsection{Independent training runs}\label{app:seeds}
Table~\ref{tab:seeds} repeats the comparison of \method{}, DAPO and the within-response top 20\% window with two more data seeds each; the data seed sets the order in which the training prompts are drawn. On both AIME sets each of the six runs with a within-response window is above each of the three DAPO runs, and on every set the margin between \method{} and DAPO in Table~\ref{tab:main} is the median of the three seeds. Paired by seed, the difference between \method{} and DAPO is $+0.077\pm0.122$ / $+0.105\pm0.088$ / $+0.091\pm0.062$ on MATH500 / AIME24 / AIME25 (mean $\pm$ standard deviation over the three seeds); between \method{} and the top 20\% window it is $+0.043\pm0.059$ / $-0.021\pm0.017$ / $-0.003\pm0.011$. Table~\ref{tab:expvstop} compares the completion, length and training tokens of the two windows over the same seeds.

\begin{table}[t]\centering\footnotesize
\caption{Three data seeds: late averages (mean@32), the differences between runs of the same seed, and the mean $\pm$ standard deviation of each row over the seeds. Differences, means and standard deviations are computed before rounding.}
\label{tab:seeds}
\setlength{\tabcolsep}{5pt}
\begin{tabular}{llcccc}\toprule
Set & Run & seed 201 & seed 300 & seed 400 & mean $\pm$ s.d. \\ \midrule
MATH500 & \method{} & 0.876 & 0.831 & 0.872 & $0.860\pm0.025$ \\
 & DAPO & 0.830 & 0.858 & 0.660 & $0.783\pm0.107$ \\
 & top 20\% window & 0.854 & 0.722 & 0.877 & $0.817\pm0.084$ \\
 & \method{} $-$ DAPO & +0.046 & $-$0.027 & +0.212 & $+0.077\pm0.122$ \\
 & \method{} $-$ top 20\% window & +0.022 & +0.110 & $-$0.004 & $+0.043\pm0.059$ \\ \midrule
AIME24 & \method{} & 0.338 & 0.278 & 0.360 & $0.326\pm0.042$ \\
 & DAPO & 0.244 & 0.256 & 0.163 & $0.221\pm0.050$ \\
 & top 20\% window & 0.380 & 0.290 & 0.372 & $0.347\pm0.050$ \\
 & \method{} $-$ DAPO & +0.094 & +0.023 & +0.197 & $+0.105\pm0.088$ \\
 & \method{} $-$ top 20\% window & $-$0.041 & $-$0.011 & $-$0.011 & $-0.021\pm0.017$ \\ \midrule
AIME25 & \method{} & 0.277 & 0.217 & 0.279 & $0.258\pm0.035$ \\
 & DAPO & 0.177 & 0.192 & 0.131 & $0.167\pm0.032$ \\
 & top 20\% window & 0.280 & 0.232 & 0.272 & $0.261\pm0.026$ \\
 & \method{} $-$ DAPO & +0.100 & +0.026 & +0.148 & $+0.091\pm0.062$ \\
 & \method{} $-$ top 20\% window & $-$0.003 & $-$0.014 & +0.007 & $-0.003\pm0.011$ \\
\bottomrule
\end{tabular}
\end{table}

\makeatletter\setlength{\@fptop}{0pt}\setlength{\@fpsep}{24pt}\makeatother
\section{Related-work comparison}\label{app:related}
Table~\ref{tab:related} places \method{} among the token-level methods of Section~\ref{sec:related}.

\begin{table}[htbp]
\centering
\caption{Token reweighting and entropy control in critic-free RLVR, as each paper describes its method. Among prior methods only OAR-P
measures the fork (how much replacing the token shifts the final-answer distribution), per token and with one extra forward pass per token; \method{} measures it
offline, on the base policy and aggregated over the entropy rank, and builds its weight on the fork probability alone, with no extra pass.}
\label{tab:related}
\scriptsize\setlength{\tabcolsep}{3pt}
\begin{tabular}{@{}L{2.45cm}L{2.85cm}L{1.0cm}L{2.35cm}L{1.8cm}L{2.3cm}@{}}
\toprule
Method & Forking proxy & Hard/soft & Fixed/adaptive & Fork measured? & Entropy control \\
\midrule
\multicolumn{6}{@{}l}{\emph{Token reweighting}} \\
80/20 mask$^\dagger$~\citep{wang2025_8020} & token entropy, batch top 20\% & hard & fixed fraction & no & none \\
GTPO / GRPO-S~\citep{gtpo2025} & entropy share & soft & fixed rule & no & none \\
Cov.-aware GRPO~\citep{taming2026} & probability--advantage covariance & soft & data-dependent kernel & no & stable entropy (stated aim) \\
OAR-P~\citep{oar2026} & counterfactual perturbation & soft & fixed rule & yes, per token (one forward pass per token) & none \\
OAR-G~\citep{oar2026} & input-gradient saliency & soft & fixed rule & no & none \\
IAPO~\citep{iapo2026} & conditional MI with the answer & soft & fixed rule & no & none \\
SC-GRPO$^\dagger$~\citep{scgrpo2026} & self-conditioned per-token KL & soft & fixed rule & no & none \\
\midrule
\multicolumn{6}{@{}l}{\emph{Entropy control}} \\
Clip-higher / clip-low~\citep{yu2025dapo,cliplow2025} & none & --- & --- & no & asymmetric trust region \\
Clip-Cov / KL-Cov~\citep{cui2025entropy} & covariance (control only) & hard / soft & fixed fraction & no & clip / penalize high-covariance tokens \\
ERPO~\citep{erpo2026} & entropy (decision pivots) & soft & adaptive gate & no & entropy-aware gating \\
STARE$^\dagger$~\citep{stare2026} & batch surprisal quantiles & piecewise & adaptive, closed loop & no & target-entropy loop \\
\midrule
\method{} (ours) & fork-probability envelope over the within-response entropy rank & soft & fixed widths; class masses equalized by the floor & offline: base policy, aggregated over the rank & entropy-gated fallback to uniform credit \\
\method{}-D (ours) & same & soft & widths tracked per batch; class masses equalized by rescaling & same & same \\
\bottomrule
\end{tabular}

\smallskip
\parbox{\linewidth}{\scriptsize\raggedright Entropy control: what a method adds to its host recipe; every run keeps its host's upper clip ratio (Tables~\ref{tab:arms} and~\ref{tab:baselines}). Extra cost per response: OAR-P adds a forward pass per token, SC-GRPO a teacher forward pass and OAR-G a saliency pass. $^\dagger$Reproduced and compared in Table~\ref{tab:main}.}
\end{table}

\section{Local displacement of the policy}\label{app:disp}
This appendix gives the protocol and every measurement (Table~\ref{tab:disp_full}) behind the local displacement of
Section~\ref{sec:entdyn} (Figure~\ref{fig:dyn}).

\paragraph{Protocol.}
The states are three checkpoints of the \method{} run, steps 250, 275 and 300 (batch entropies $1.90$, $3.24$ and $3.25$ nats), and, as a
control, the step-300 checkpoint of the DAPO run ($4.10$ nats). From each checkpoint two continuations of five steps are trained with the
checkpoint's optimizer state and data order: one keeps the window with the fallback switched off ($m_{\mathrm{gov}}\equiv1$), the other uses
uniform credit ($m_{\mathrm{gov}}\equiv0$, the damping $(1-\hat p_i)^\gamma$ kept); damping, clipping, dynamic sampling and every other setting are those of training. At step 275 a third
continuation freezes the fallback near $m_{\mathrm{gov}}=0.5$ (band $1.24/5.24$ nats, realized $m_{\mathrm{gov}}=0.49$--$0.57$ over the five steps).
For the measurement, 128 training prompts $\times$ 16 rollouts are sampled from the starting policy with the training sampler (temperature
$1.0$, top-$p$ $1.0$, 20{,}480-token cap), and a teacher-forced pass under the starting policy and under each continued policy records the sampled
token's log-probability and the conditional entropy at every response position. The displacement is the per-token mean of
$\log\pi_t(x_t\mid s_t)-\log\pi_{t+5}(x_t\mid s_t)$, the estimator $k_1$, which is unbiased for
$\mathbb E_{s\sim\pi_t}\mathrm{KL}(\pi_t(\cdot\mid s)\,\|\,\pi_{t+5}(\cdot\mid s))$ on the starting policy's samples; $k_3=r-1-\log r$ with
$r=\pi_{t+5}/\pi_t$ is reported as a second estimator. Intervals are a bootstrap over prompts (2{,}000 draws); the class split uses the verifier outcome of the
rollout; $\Delta H$ is the per-token mean change of the conditional entropy on the same contexts. The parameter displacement is the Euclidean
norm of $\theta_{t+5}-\theta_t$ over all $8.19$B actor parameters. The uniform continuation at step 275 was trained twice; the two runs agree
within $1\%$ in KL and $2\%$ in parameter displacement.

\begin{table}[htbp]\centering\scriptsize
\caption{Every continuation of the displacement measurement: parameter displacement $\|\theta_{t+5}-\theta_t\|_2$, KL per token from the
starting policy (in units of $10^{-3}$; $k_1$ with its 95\% prompt bootstrap, $k_3$), change of the conditional entropy on the starting contexts
(nats per token) and the KL split by the rollout's class (correct / wrong, same units); $\bar H$: token-mean entropy of the starting batch (nats).}
\label{tab:disp_full}
\setlength{\tabcolsep}{3pt}
\begin{tabular}{llllllll}\toprule
State & $\bar H$ & continuation & $\|\Delta\theta\|$ & KL $k_1$ [95\%] & $k_3$ & $\Delta H$ & KL correct / wrong \\ \midrule
\DispFullRows
\bottomrule
\end{tabular}
\end{table}

\paragraph{Results.}
At step 300 the gradient norm of the uniform continuation exceeds the clipping norm at every step and that of the windowed continuation in four of the five, which equalizes their parameter steps (ratio $0.99$, Figure~\ref{fig:dyn}c). Per unit of squared parameter displacement the windowed update moves the policy
$2$--$7$ times further at the three \method{} states and $1.4$ times further at the DAPO-trained control. There the KL ratio itself is $0.86$, because the windowed continuation takes the smaller parameter step. At step 275 the frozen-fallback continuation lies between the two in both
quantities, $3\%$ from the linear interpolation in $m_{\mathrm{gov}}$. At step 250 the batch entropy of the continuations separates within five
steps (window kept $2.08\to3.27$ nats, uniform $2.07\to2.58$) while the conditional entropy on the fixed starting contexts rises slightly under
both ($+0.017$ nats per token under the window and $+0.077$ under uniform credit): the rise under the window comes from the change of the generated contexts. The excess
displacement of the windowed update is spread over the whole response: its share in the top 20\% of the entropy rank is $0.24$--$0.36$ against
$0.29$--$0.37$ for the uniform update, and its ratio to the uniform update is largest in the lower half of the rank at steps 250 and 300.

\end{document}